\documentclass[runningheads]{llncs}
\usepackage[T1]{fontenc}
\usepackage{graphicx}
\usepackage[utf8]{inputenc} 
\usepackage[T1]{fontenc}    
\usepackage{booktabs}       
\usepackage{amsfonts}       
\usepackage{nicefrac}       
\usepackage{microtype}      
\usepackage{xcolor}         
\usepackage{amsmath}
\usepackage{graphicx}
\usepackage{subcaption}
\usepackage{adjustbox}
\usepackage{hyperref}
\usepackage{cleveref}
\Crefname{equation}{Eq.}{Eqs.}
\Crefname{figure}{Fig.}{Figs.}
\Crefname{table}{Tab.}{Tabs.}
\Crefname{section}{Sec.}{Secs.}

\crefname{equation}{eq.}{eqs.}
\crefname{figure}{fig.}{figs.}
\crefname{table}{tab.}{tabs.}
\crefname{section}{sec.}{secs.}
\usepackage{wrapfig}
\usepackage{tikz}

\usepackage{tablefootnote}  
\usepackage{array} 
\usepackage{multirow}  
\newcolumntype{M}[1]{>{\centering\arraybackslash}m{#1}}

\begin{document}
\title{CURE: Controllable Unified Image Restoration for Complex Degradations}

%

%
\author{Boseong Kim \and Donghyeon Cho\thanks{Corresponding author.}}
\institute{Department of Computer Science, Hanyang University, Seoul, South Korea \\
\email{\{hosky789, doncho\}@hanyang.ac.kr}}

\maketitle              
\begin{abstract}
The presence of composite degradations poses a significant challenge, since the underlying corruption factors exhibit complex and interdependent interactions.
%
%
Even when the degradation types are known, accurately restoring the image remains difficult due to the intertwined nature of their effects and the need for selective control during the recovery process.
To address this, we introduce CURE, a unified framework that enables controllable restoration in complex degradation settings by learning disentangled and adjustable representations.
CURE is driven by four complementary objectives.
First, an identity embedding is incorporated, along with a reconstruction constraint, to ensure that the model can reproduce the input image when restoration is unnecessary.
Second, the ratio control mechanism blends the identity embedding with degradation-specific embeddings using user-regulated mixing ratios, allowing continuous control over restoration intensity.
Third, an intermediate loss is applied to supervise stepwise outputs, each encouraged to tackle the removal of only a single degradation factor within a composite mixture.
Finally, a permutation-invariant loss ensures that the model achieves consistent restoration quality regardless of the order in which multiple degradations are addressed.
Since CURE modifies only the training strategy and not the underlying network architecture, it can be seamlessly integrated into existing controllable restoration models. 
Experiments demonstrate that CURE delivers state-of-the-art performance on composite degradation benchmarks, while enabling both selective and jointly fused restoration through flexible modulation of embedding ratios.
{The code and dataset are available at \url{https://github.com/bo-oseng/CURE}.}
\end{abstract}

\section{Introduction}
\label{sec:intro}
%
Image restoration is a core task in low-level vision aiming to recover high-quality images from degraded inputs.
This capability is important in real-world applications such as autonomous driving~\cite{lin2025jarvisir} and robotics~\cite{porav2019can}, where obtaining clean and stable visual signals is often difficult.
The problem remains inherently ill-posed because multiple plausible clean images can explain the same degraded content, which makes it difficult for models to generalize across diverse degradation patterns.
Earlier approaches~\cite{he2010single,qu2017deshadownet,zhang2018density} rely on hand-crafted priors designed for specific degradation scenarios.
With the introduction of transformer architectures~\cite{vaswani2017attention}, attention-based models such as Uformer~\cite{RN21} and Restormer~\cite{RN25} show strong performance improvements.
However, these models are typically optimized for a single degradation type, requiring separate networks for different corruptions and limiting their flexibility when dealing with varied or mixed degradation conditions.

To address the limitations of single-task models, the all-in-one restoration paradigm aims to build a unified network capable of handling a broad spectrum of restoration tasks.
Existing attempts include corruption-agnostic approaches, such as AirNet~\cite{RN29}, and weather-oriented architectures, such as WGWS~\cite{RN27} and TKL~\cite{RN5}.
Beyond these approaches, recent works aim to enhance task-discriminative capability through advanced learning objectives, such as mixture-of-experts and degradation classification~\cite{zamfir2025complexity,hu2025universal}.
%
From a different perspective, prompt-based methods, including PromptIR~\cite{RN9}, AM-PromptIR~\cite{zhang2025adaptive}, TextPromptIR~\cite{yan2025textual}, VLU-Net~\cite{zeng2025vision}, and DFPIR~\cite{tian2025degradation} further extend this direction by using learned embeddings to guide the model’s behavior in a task-adaptive manner.
Despite these developments, most studies still treat individual degradation types independently and do not fully address scenarios where multiple corruptions coexist in complex combinations.

Recently, composite models such as OneRestore~\cite{RN34} have emerged to address simultaneous degradations. 
Nevertheless, they still struggle with clean disentanglement and controllability.
While OneRestore represents meaningful progress, it still encounters difficulties in cleanly disentangling and removing mixed degradations.
A key issue lies in its limited controllability.
(1) The model does not provide an explicit mechanism to bypass restoration and preserve the input content when desired.
While a "$clear$" prompt can be used to mimic an identity mapping, existing models do not allow users to intentionally preserve certain degradations or bypass the processing for specific components.
(2) The model also lacks support for soft or partial removal.
For instance, when a user prefers to retain a mild level of haze for stylistic purposes, the model offers no control over the degree of removal.
These controllability limitations become even more problematic under composite degradations.
(3) Selectively removing a single corruption within a mixture often introduces unintended artifacts.
(4) Moreover, the restoration outcome is sensitive to the sequence in which degradations are processed; for example, removing snow before enhancing low-light yields noticeably different results from performing these steps in the opposite order.
Such issues highlight the fundamental need for more explicit and reliable controllability when restoring images affected by multiple degradations.

To address these challenges and strengthen controllability in composite degradation settings, we propose Controllable Unified Image Restoration (CURE), a disentangled restoration framework that improves the adaptability and effectiveness of all-in-one models.
CURE integrates several components that provide structured and user-controllable restoration:
(1) Identity embedding and an identity loss, which constrain the model to preserve the input when no restoration is required.
(2) A ratio-control loss, which enables proportional removal of degradations by mixing the identity embedding with degradation-specific embeddings at a controllable ratio.
(3) An intermediate loss, which supervises partially restored outputs and guides the model to learn fine-grained restoration behaviors.
(4) A permutation-invariant loss, which ensures consistent results even when degradations are addressed in different orders.
Together, these components form a disentangled restoration pipeline that offers flexible, structured, and adaptive control across various degradation combinations.
CURE can remove multiple degradations either step-by-step or in a single pass, while also allowing users to specify the desired removal intensity for each degradation.
%
Extensive experiments on composite degradation benchmarks demonstrate that CURE achieves state-of-the-art restoration performance while offering significantly enhanced controllability and robustness compared to existing frameworks.

\section{Related work}
\label{sec:related_work}

\noindent \textbf{Single Task Image Restoration.}
%
Image restoration encompasses various low-level vision tasks, including dehazing~\cite{He2009,zheng2023curricular}, deraining~\cite{Fu2017,Chen2024}, desnowing~\cite{Bossu2011,quan2023image}, and low-light enhancement~\cite{Land1977,zhou2023low}.
While the aforementioned methods perform well on individual or at most two degradations, they require separate models for different tasks. 
This paper explores a unified model that handles multiple degradations simultaneously while allowing selective removal of specific ones.

\noindent \textbf{All-in-One Image Restoration.}
Recent developments in image restoration have explored various techniques to handle images affected by multiple types of degradation. 
One prominent approach is the partial parameter-sharing One-to-Many strategy, which utilizes a shared backbone with multiple input and output pathways to address different degradation factors~\cite{RN15,Chen2021,Wang2023}. 
Beyond this, recent efforts have shifted toward fully shared-weight architectures for universal restoration, enabling more flexible solutions.
For instance, TKL~\cite{RN5} proposed a unified model for adverse weather removal using a two-stage knowledge-learning framework, while AirNet~\cite{RN29} introduced a model that handles unknown corruptions without prior knowledge.
%
%
To improve performance, MoCE-IR~\cite{zamfir2025complexity} employs mixture-of-experts to discriminate tasks based on complexity, while DCPT~\cite{hu2025universal} utilizes degradation classification to explicitly guide the restoration process.
%
%
{
Meanwhile, recent methods like and UniRestore~\cite{chen2025unirestore} leverage diffusion models, while MambaIR~\cite{guo2024mambair} utilizes state-space models for universal restoration.
While these approaches primarily focus on novel network structures, CURE introduces a model-agnostic training paradigm that can complement such structural innovations.
}

\noindent \textbf{Prompt-based Approaches.}
Since existing universal methods struggle with interference between different degradation types in complex scenarios, prompt-based approaches are emerging to adaptively guide restoration across multiple degradations.
PromptIR~\cite{RN9} is a prompt-based learning approach for all-in-one image restoration that encodes degradation-specific information to dynamically guide restoration.
To further advance this concept, text-guided restoration approaches~\cite{conde2024instructir,zhang2025adaptive,yan2025textual,zeng2025vision} take advantage of text embeddings and language models to provide more precise guidance.
Specifically, VLU-Net~\cite{zeng2025vision} addresses the manual tuning constraints of deep unfolding networks using text-driven gradients, while preserving interpretability.
On the other hand, AM-PromptIR~\cite{zhang2025adaptive}, TextPromptIR~\cite{yan2025textual}, InstructIR~\cite{conde2024instructir} and DFPIR~\cite{tian2025degradation} leverage the capabilities of language models to significantly boost restoration performance.
Nevertheless, since controllable restoration is achieved implicitly rather than being treated as a core objective, these methods lack the level of precision required for effective control.
%
%

\noindent \textbf{Composite Image Restoration.}
Composite image restoration deals with images degraded by multiple factors simultaneously.
This task poses a unique challenge as the entanglement of different degradations leads to severe interference during the restoration process.
To tackle this, OneRestore~\cite{RN34} successfully introduces a text-embedding strategy utilizing versatile scene descriptors to enable adaptive restoration for such diverse composite degradations.
Despite its effectiveness in a unified context, similar to previous controllable frameworks, it lacks the capability for precise selective degradation removal and suffers from order dependency in restoration sequences.
This deficiency primarily stems from an insufficient consideration of the complex entanglement between degradations, where the blended characteristics of multiple corruptions make it difficult to isolate individual factors.
%
%
%
Addressing these limitations, we propose CURE, a controllable restoration framework that leverages text-guided embeddings to achieve fine-grained, disentangled removal of complex degradations.
%

\section{Method}
\label{sec:method}
\begin{figure*}[t]
    \begin{center}
        \def\arraystretch{0.4}
        \begin{tabular}{@{}c}    
            \includegraphics[width=0.9\linewidth]{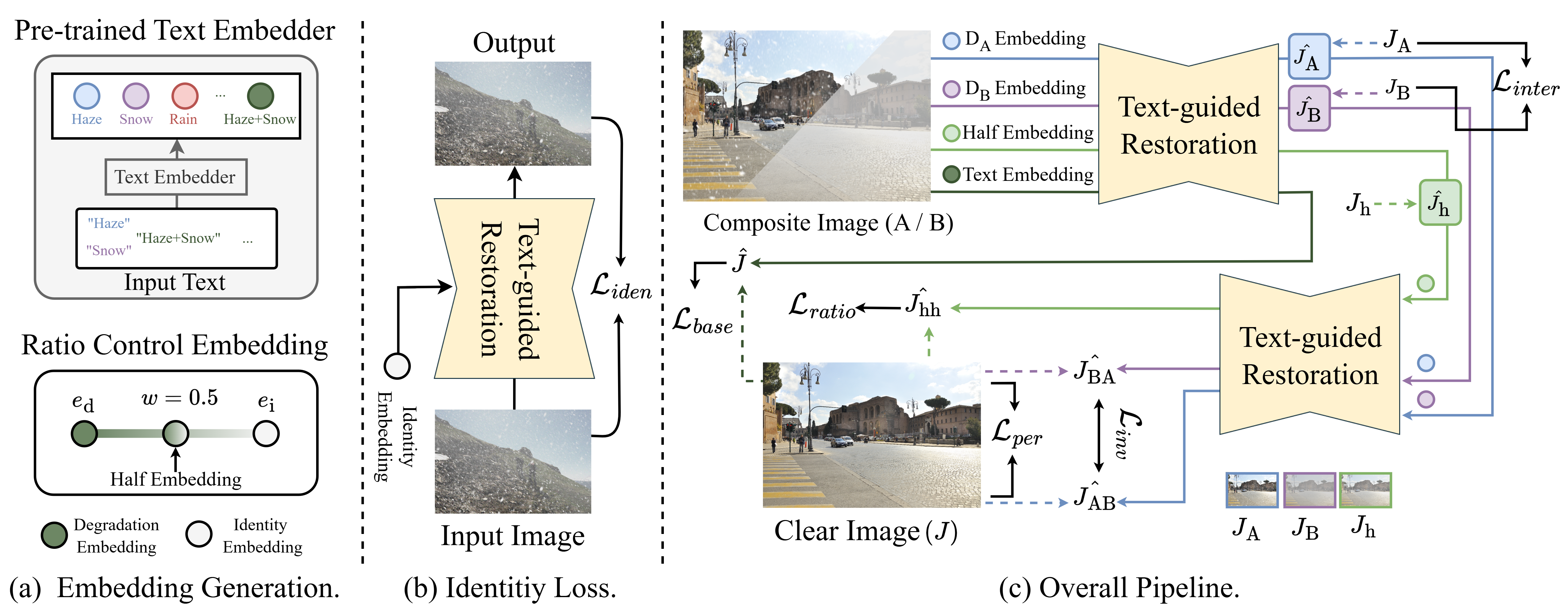}   
        \end{tabular}
    \end{center}
    \caption{\textbf{Overview of our CURE.} (a) Embedding Generation constructs degradation-aware embeddings via a pre-trained text embedder and our ratio-control mechanism. 
    This enables continuous intensity control by interpolating between identity and degradation embeddings.
    (b) The identity loss is a loss function designed specifically for the identity input, which preserves the input without any modifications. 
    It serves to suppress changes in the text-guided restoration model's output when the model receives an identity embedding as input.
    (c) Overall Pipeline of our CURE. The text-guided restoration model handles composite images with multiple degradations  ($D_{A}$ and $D_{B}$ simultaneously, denoted as $D_{AB}$) using embeddings selected from Pre-trained Text Embedder sets (a). The Half Embedding is formed by weighted combination of Identity and $D_{AB}$ embeddings, and each restoration flow is processed following the guidance of its corresponding same-colored pipeline
    }
    \label{fig:overview}
\end{figure*}

%
To decouple intertwined degradations and achieve fine-grained control, we propose a unified framework leveraging disentangled learning.
Specifically, we construct degradation-aware embeddings via a pre-trained text embedder and ratio-control mechanism, as illustrated in \Cref{fig:overview}(a).
The identity loss shown in \Cref{fig:overview}(b) preserves input content, essential for disentangled control.
Finally, these elements are unified into a restoration pipeline, as shown in \Cref{fig:overview}(c), allowing the model to adaptively handle composite degradations under precise user guidance.

\subsection{Baseline Framework}
Our approach operates within a text-guided restoration framework based on an encoder-decoder architecture.
The framework receives an image $I$ and a text prompt as inputs.
The prompt is encoded into a text embedding, which is then used to modulate the restoration process through cross-attention or feature-injection mechanisms.
Following existing text-guided restoration models, the baseline is optimized using a combination of a reconstruction objective and a task-specific auxiliary loss.
We formulate this general baseline loss as:
\begin{equation}
    \mathcal{L}_\text{base} = \mathcal{L}_\text{rec}(J, \hat{J}) + \lambda \mathcal{L}_\text{aux},
    \label{eq:baseline_loss}
\end{equation}
where $J$ and $\hat{J}$ denote the ground truth (GT) and the model output, respectively. 
In addition, $\mathcal{L}_\text{rec}$ represents the pixel-wise reconstruction loss, $\mathcal{L}_\text{aux}$ denotes the auxiliary regularization term tailored to specific model designs, and $\lambda$ is a balancing hyperparameter.
Specifically, the implementation of Eq.~\eqref{eq:baseline_loss} varies across different baseline methods.
For instance, TextPromptIR~\cite{yan2025textual} primarily relies on the ${L}_1$ distance for $\mathcal{L}_\text{rec}$ without an auxiliary term (i.e., $\lambda=0$).
%
%
In the case of OneRestore~\cite{RN34}, the reconstruction term $\mathcal{L}_\text{rec}$ combines the Smooth ${L}_\text{1}$ and MS-SSIM losses, while the auxiliary term $\mathcal{L}_\text{aux}$ is formulated as a contrastive loss utilizing negative degradation samples.
In this work, we retain the inherent $\mathcal{L}_\text{base}$ of the selected baseline to preserve its foundational restoration capabilities, while introducing our proposed objectives to enable disentangled control.

\subsection{Identity Embedding}
\label{stage:identity}
The identity operation refers to the case where the model outputs the input image without any modifications.
While baseline text embedders are designed for restoration tasks, they do not account for the identity operation.
To support this, we assign an embedding vector that explicitly represents the identity operation.
This embedding vector is predefined as a constant vector of ones and requires no additional training.
We refer to this constant vector as the identity embedding.
The baseline restoration models are trained to preserve the input image unchanged when the identity embedding is fused into their text-guided frameworks.

\noindent \textbf{Ratio-Control Embedding.}
Existing frameworks are designed to fully restore a clean image regardless of the severity of degradation, as long as a corresponding degradation embedding is provided.
This observation suggests the potential to control the degree of restoration by adjusting the degradation embedding vector.
To enable this, we leverage the identity embedding that preserves the input without any modification.
Specifically, we linearly interpolate between the identity embedding $\mathbf{e}_{\text{i}}$ and the degradation embedding $\mathbf{e}_{\text{d}}$ according to the desired restoration intensity ratio, as follows: 
%
\begin{equation}
    \mathbf{e}_{\text{r}} = (1 - w)\cdot\mathbf{e}_{\text{i}} + w \cdot \mathbf{e}_{\text{d}},
    \label{eq:intensity_control_embedding}
\end{equation}
%
where $w$ is an intensity ratio control parameter ranging from identity operation ($w = 0$) to full restoration ($w = 1$), and $\mathbf{e}_{\text{r}}$ denotes the corresponding ratio-control embedding.
Adjusting $w$ enables the model to continuously control the restoration intensity for each degradation type, allowing for proportional rather than complete removal of degradations.
Note that the ratio-control embedding, like other degradation text embeddings, is integrated into the text-guided modules of the baseline models.
This allows it to explicitly guide the network to remove degradations proportionally to the specified ratio.
\subsection{Loss Functions}
Based on the identity embedding and ratio-control embedding, CURE is trained through disentangled learning guided by the following four loss functions.

\noindent \textbf{Identity Loss.}
The identity loss is designed to ensure that the model preserves the original content when no degradation needs to be removed.
In other words, it enforces the model to maintain the input unchanged when the identity embedding is used as a prompt in the restoration model, as follows:
\begin{equation}
    \mathcal{L_\text{iden}} = \mathcal{L_\text{rec}}(I,\hat{I}),
    \label{eq:identity_loss}
\end{equation}
%
where ${I}$ is the input image, $\hat{I}$ represents the model output with identity embedding.
Through this loss term, the model learns to behave as an identity function when necessary, which can be utilized for the intermediate and permutation-invariant losses in cases involving a single degradation type.

\noindent \textbf{Ratio-Control Loss.}
The goal of the ratio-control loss is to ensure that, when the ratio-control embedding is used as a prompt, the model removes degradations proportionally to the specified ratio, rather than performing full restoration.
Ideally, this would require GT images corresponding to each value of $w$ in \Cref{eq:intensity_control_embedding}, where only the $w$ proportion of degradation is removed.
However, in practice, constructing GT images for every possible $w$ is infeasible. 
Therefore, during training, we only generate and use GT images corresponding to $w = 0.5$.
In the experiments, we demonstrate that training with only the $w = 0.5$ case generalizes well, and the model performs seamlessly for other $w$ values during inference.
Accordingly, the ratio-control loss is defined using only the half-intensity case during training, as follows:
%
\begin{equation}
    \mathcal{L}_{\text{ratio}} = \mathcal{L_\text{rec}}(J_\text{h},\hat{J_\text{h}}) + \mathcal{L_\text{rec}}(J,\hat{J_\text{hh}}),
    \label{eq:intensity_control_loss}
\end{equation}
%
where $J_{\text{h}}$ denotes the half-degraded GT image, $\hat{J}_{\text{h}}$ is the corresponding half-restored output, $J$ is the fully clean GT image, and $\hat{J}_{\text{hh}}$ is the fully restored output obtained by reapplying the restoration model to $\hat{J}_{\text{h}}$.
Note that both $\hat{J}_{\text{h}}$ and $\hat{J}_{\text{hh}}$ are generated using the ratio-control embedding for $w = 0.5$.
The first term supervises the half-restored output $\hat{J}_{\text{h}}$ to match the corresponding half-degraded GT $J_{\text{h}}$.
This explicitly trains the model to remove approximately half of the degradation, establishing fine-grained controllability as intended by the ratio-control embedding.
The second term applies the same half-intensity embedding again, guiding the final output $\hat{J}_{\text{hh}}$ to progressively approach the fully clean image $J$.
This enforces a self-consistency constraint that reinforces the linearity of the control space.

\noindent \textbf{Intermediate Loss.}
To improve controllability and prevent error propagation in sequential restoration, we propose an intermediate loss function that provides precise and explicit supervision for removing each specific degradation at its respective stage in the restoration process.
Let $D_{A}$ and $D_{B}$ denote two independent degradations present in a composite image.
If the image is restored sequentially by first removing $D_{A}$ and then $D_{B}$, or in reverse order, any errors introduced in the first step are likely to propagate into the second.
To mitigate this issue, the intermediate loss is designed to help the model handle each degradation independently during the restoration process, as follows:
%
\begin{equation}
    \mathcal{L_\text{inter}} = \mathcal{L_\text{rec}}(J_\text{A},\hat{J_\text{A}}) + \mathcal{L_\text{rec}}(J_\text{B},\hat{J_\text{B}}),
    \label{eq:intermediate_loss}
\end{equation}
%
where $J_\text{A}$ denotes the GT image with degradation $D_{B}$ removed (but not completely clean), and $\hat{J}_\text{A}$ is the model output when removing only $D_{B}$.
Similarly, $J_\text{B}$ is the GT image with degradation $D_{A}$ removed, and $\hat{J}_\text{B}$ is the corresponding output when the model restores only $D_{A}$.
%
%
Note that~\Cref{eq:intermediate_loss} is not limited to dual degradations and can be seamlessly applied to multiple types.
Also, for single-degradation scenarios, one of $D_{A}$ or $D_{B}$ can be treated as “clear,” and the identity embedding can be applied accordingly.

\noindent \textbf{Permutation-Invariant Loss.}
To ensure consistency in sequential restoration of composite degradations, we propose a permutation-invariant loss that enforces the final output to remain unaffected by the order in which degradations are removed.
To ensure consistent results regardless of the restoration order, our permutation-invariant loss consists of two terms as follows:
%
\begin{equation}
    \mathcal{L_\text{per}} = \mathcal{L_\text{rec}}(J,\hat{J_\text{AB}}) + \mathcal{L_\text{rec}} (J,\hat{J_\text{BA}}), \quad
    \mathcal{L_\text{inv}} = \mathcal{L}_{L1}(\hat{J_\text{AB}},\hat{J_\text{BA}}),
    \label{eq:permutation-invariant}
\end{equation}
%
where $\hat{J}_\text{AB}$ and $\hat{J}_\text{BA}$ represent the model outputs obtained by restoring the image in the order of $D_{A}$ followed by $D_{B}$, and in the reversed order.
The $\mathcal{L}_\text{per}$ ensures that both restoration sequences produce outputs close to the $J$.
The invariance loss $\mathcal{L}_\text{inv}$, calculated via $L_1$ or Smooth $L_1$ distance depending on the baseline, enforces consistency between the two restoration outputs, $\hat{J}_\text{AB}$ and $\hat{J}_\text{BA}$.
Finally, the restoration model is trained using the baseline loss~\Cref{eq:baseline_loss}, along with the four proposed loss functions: identity~\Cref{eq:identity_loss}, ratio-control~\Cref{eq:intensity_control_loss}, intermediate~\Cref{eq:intermediate_loss}, and permutation-invariant loss functions~\Cref{eq:permutation-invariant}.

%
%
\begin{table*}[t]
    \centering
    \caption{Quantitative comparisons on CCDD-11 dataset\tablefootnote{The performance gap compared to the originally reported OneRestore~\cite{RN34} results is primarily due to modifications made to all rain-related datasets. Unlike the original setup, which uses a limited variety of rain masks in the CDD-11 dataset, our version incorporates a broader and more diverse set of rain mask variations.}.}
    \label{tb:comparison}
    \resizebox{0.8\linewidth}{!}
    {   
        \begin{tabular}{l|c|c|c|c}
            \hline
            Types & Methods & PSNR $\uparrow$ & SSIM $\uparrow$ & \#Params \\
            \hline \hline
            & Input & 16.04 & 0.61 & - \\
            \hline \hline
            One-to-One
            & MIRNet~\cite{zamir2020learning} & 26.17 & 0.87 & 31.79M \\
            & MPRNet~\cite{RN8} & 27.07 & 0.86 & 15.74M \\
            & MIRNetV2~\cite{zamir2022learning} & 24.90 & 0.83 & 15.74M \\
            & Restormer~\cite{RN25} & 25.43 & 0.84 & 26.13M \\
            & DGUNet~\cite{mou2022deep} & 27.17 & 0.86 & 17.33M \\
            & NAFNet~\cite{chen2022simple} & 26.78 & 0.80 & 17.11M \\
            & Fourmer~\cite{zhou2023fourmer} & 24.49 & 0.82 & 0.55M \\ 
            & OKNet~\cite{cui2024omni} & 27.54 & 0.87 & 4.72M \\
            \hline\hline
            One-to-Many
            & AirNet~\cite{RN29} & 26.33 & 0.84 & 8.93M \\
            & WGWSNet-WG~\cite{RN27} & 25.40 & 0.85 & 25.76M \\
            & WGWSNet-WS~\cite{RN27} & 20.20 & 0.74 & 25.76M \\
            & PromptIR~\cite{RN9} & 28.07 & 0.87 & 38.45M \\
            & AdaIR~\cite{cui2025adair} & 27.25 & 0.86 & 28.77M \\
            & DFPIR ~\cite{tian2025degradation} & 27.51 & 0.85 & 31.10M \\
            \hline\hline
            Text-guided
            & AM-PromptIR~\cite{zhang2025adaptive} & 26.11 & 0.84 & 64.54M \\ 
            & TextPromptIR~\cite{yan2025textual} & 27.55 & 0.86 & 93.55M \\ 
            & OneRestore (Visual)~\cite{RN34} & 27.38 & 0.86 & 5.98M \\ 
            & OneRestore (Text)~\cite{RN34} & 27.74 & 0.87 & 5.98M \\
            \hline
            Fine Control
            & AM-PromptIR$+$ CURE & 26.79 & 0.84 & 64.54M \\ 
            & TextPromptIR $+$ CURE & 28.15 & 0.87 & 93.55M \\ 
            & OneRestore $+$ CURE (Visual) & 27.89 & {0.86} & 5.98M \\ 
            & OneRestore $+$ CURE (Text) & \textbf{28.28} & 0.87 & 5.98M \\
            \hline
        \end{tabular}%
    
    }
\end{table*}
%
\begin{figure*}[t]
    \begin{center}
        \def\arraystretch{0.4}
        \begin{tabular}{@{}c}    
            \includegraphics[width=1.0\linewidth]{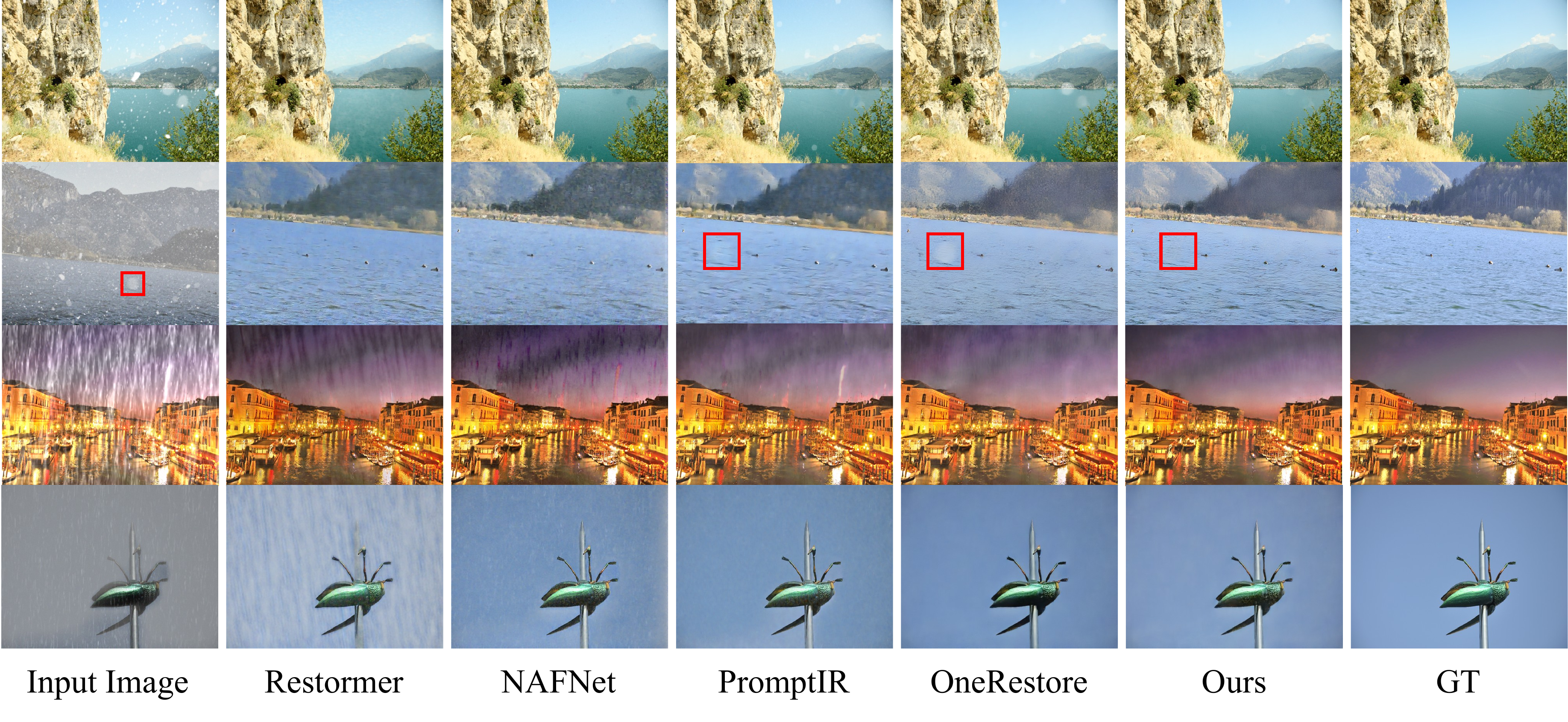}
        \end{tabular}
    \end{center}
    \caption{
    Qualitative comparisons on the CCDD-11 dataset. 
    Ours refers to the results of OneRestore integrated with our CURE.
    %
    %
    Arranged from top to bottom, the rows illustrate cases in which the input images are degraded by $snow$, $low + haze + snow$, $rain$, and $low + haze + rain$, respectively.
    %
    }
    \label{fig:comparison}
\end{figure*}
%
\begin{table}[t]
    \centering
    \caption{Ablation study on different loss functions.}
    \label{tab:losses_ablation}
    \resizebox{0.8\linewidth}{!}{
        \begin{tabular}{M{1.5cm}| M{2.0cm}| M{2.3cm}| M{3.3cm}| M{1.5cm}}
            \hline 
            Identity & Ratio control & Intermediate & Permutation-invariant & PSNR \\
            \hline \hline
            \checkmark & &  &   & 27.85 \\
            \hline
            \checkmark &\checkmark &  &   & 27.89  \\
            \hline
            \checkmark & & \checkmark &  &  28.13 \\
             \hline
            \checkmark & &  &  \checkmark &  28.20 \\
            \hline
            \checkmark & \checkmark &  \checkmark & \checkmark & \textbf{28.28} \\
            \hline
        \end{tabular}
    }
\end{table}
%

\begin{table}[t]
    \centering
    \caption{PSNR with identity embedding ($clear$ prompt for OneRestore).}
    \label{tab:identity_psnr}
    \resizebox{0.95\linewidth}{!}{
    \begin{tabular}{c|c|c|c|c|c|c}
        \hline
        Method & Low & Haze & Rain & Snow & Low$+$Haze & Low$+$Rain \\
        \hline \hline
        OneRestore~\cite{RN34} & 21.31 & 23.87 & 25.04 & 24.94 & 24.45 & 23.89  \\
        Ours       & 55.27 & 57.31 & 57.35 & 57.44 & 57.58 & 57.24  \\
        \hline \hline
        Method & Low$+$Snow & Haze$+$Rain & Haze$+$Snow & Low$+$Haze$+$Rain & Low$+$Haze$+$Snow & Average \\
        \hline \hline
        OneRestore~\cite{RN34} & 23.34 & 23.60 & 23.80 & 23.72 & 23.67 & 23.78  \\
        Ours       & 57.00 & 57.32 & 57.59 & 57.66 & 57.73 & \textbf{57.05}  \\
        \hline
    \end{tabular}}
\end{table}

%
\begin{table}[t]
        \centering
        \caption{Intensity ratio and Embedder classification accuracy.}
        \label{tab:ratio-control-accuracy}
        \resizebox{0.95\linewidth}{!}{
        \begin{tabular}{c|c|c|c|c|c|c|c|c|c|c|c}
            \hline
            Intensity Ratio $w$ & 1.0 & 0.9 & 0.8 & 0.7 & 0.6 & 0.5 & 0.4 & 0.3 & 0.2 & 0.1 & 0.0 \\
            \hline \hline
            Accuracy & 0.3 \% & 0.4 \% & 0.8 \% & 2.5 \% & 16.4 \% & 43.8 \% & 56.2 \% & 77.9 \% & 94.7 \% & 95.0 \% & 94.9 \% \\
            \hline
        \end{tabular}}
        \label{tab:intensity-ratio-accuracy}
\end{table}
%
\begin{figure}[t]
\centering
    \def\arraystretch{0.4}
    \begin{tabular}{@{}c}
        \includegraphics[width=1\linewidth]{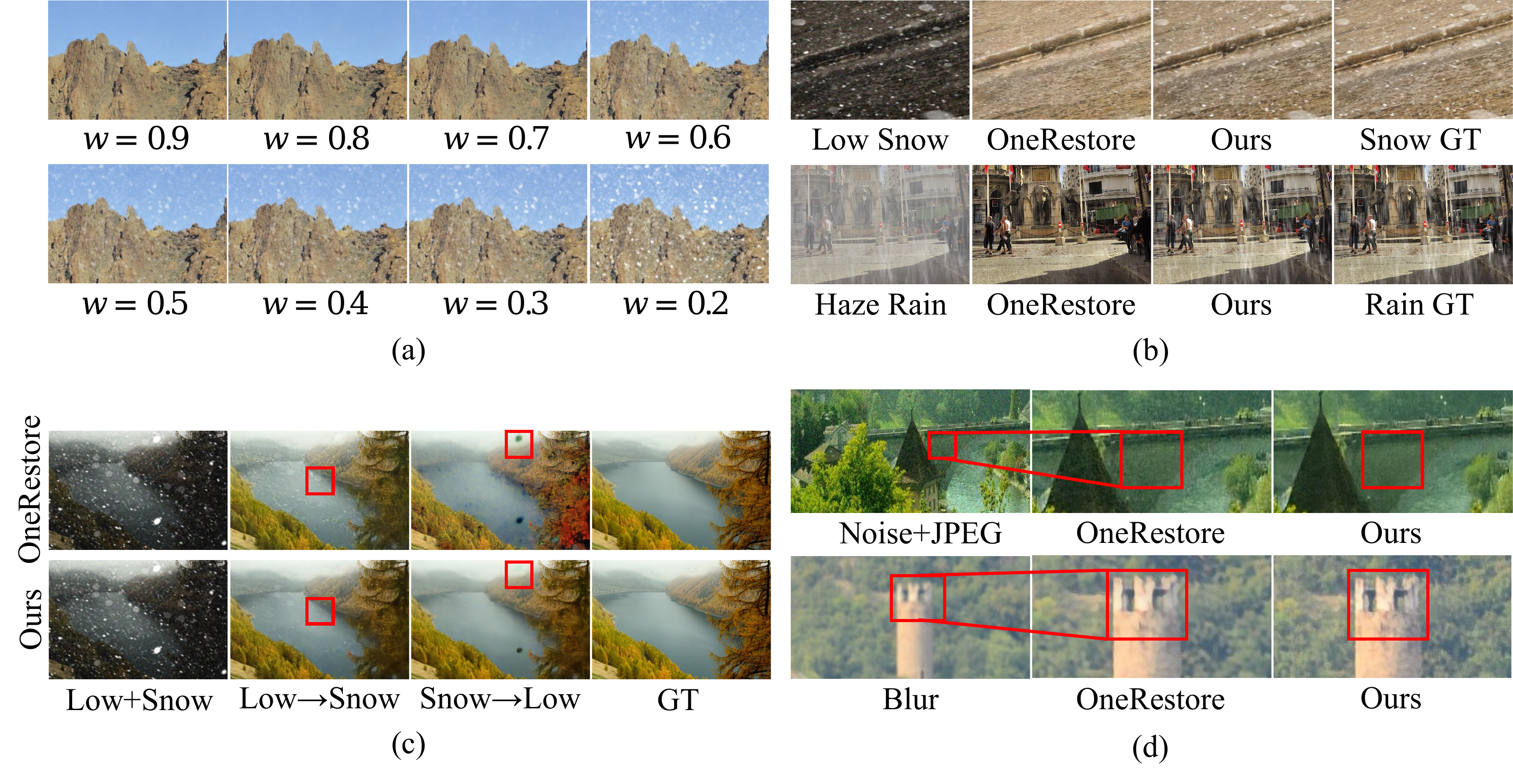}
    \end{tabular}
    \caption{(a) Ratio-control embedding. (b) Selective restoration. (c) Restoration order dependency. (d) Blur-noise-JPEG composite degradation.}
    \label{fig:qualitative-results}
\end{figure}
%
\begin{table}[t]
    \centering
    \caption{PSNR for selective restoration on two combined degradations}
    \label{tab:partial_two_composite_psnr}
    \resizebox{0.7\linewidth}{!}{
    \begin{tabular}{c||c|c|c||c|c|c}
        \hline
        Degradation & Task & OneRestore~\cite{RN34} & +CURE & Task & OneRestore & +CURE \\
        \hline \hline
        Low$+$Haze & Delow & 20.89 & 30.61 & Dehaze & 22.56 & 35.50\\
         Low$+$Rain &  Delow & 24.67 & 26.74 & Derain & 28.76 & 36.37 \\
         Low$+$Snow &  Delow & 23.42 & 26.68 & Desnow & 27.53 & 34.30\\
         Haze$+$Rain &  Dehaze & 27.73 & 33.01 & Derain & 34.69 & 40.30 \\
         Haze$+$Snow &  Dehaze & 25.56 & 33.13 & Desnow & 31.61 & 37.81\\
        \cline{1-7}
        Average & - & 24.45 & \textbf{30.03} & - & 29.03 & \textbf{36.86} \\
        \hline
    \end{tabular}
    }
\end{table}
%
\begin{table}[t]
        \centering
        \caption{Classification performance after selective restoration. The top-row degradation types represent the remaining degradations after selective restoration.}
        \label{tab:classification}
        \resizebox{0.95\linewidth}{!}{
        \begin{tabular}{c|c|c|c|c|c|c|c|c|c|c}
            \hline
            Method & Low & Haze & Rain & Snow & Low$+$Haze & Low$+$Rain & Low$+$Snow & Haze$+$Rain & Haze$+$Snow & Average \\
            \hline \hline
            OneRestore~\cite{RN34} & 76.7\% & 61.7\% & 73.8\% & 48.6\% & 76.7\% & 42.0\% & 24.5\% & 60.5\% & 15.5\% & 53.33\% \\
            $+$ CURE       & 98.6\% & 85.5\% & 93.7\% & 94.3\% & 88.7\% & 91.6\% & 98.4\% & 95.6\% & 94.6\% & \textbf{93.44\%}\\
            \hline
        \end{tabular}}
\end{table}
%
\begin{table}[t]
    \centering
    \caption{PSNR evaluation with varying restoration orders.}
    \label{tab:order-dependent}
    \resizebox{0.99\linewidth}{!}{
        \begin{tabular}{c || r@{\;}c@{\;}l | c | c || r@{\;}c@{\;}l | c | c || r@{\;}c@{\;}l | c | c}
            \hline
            Degradation & \multicolumn{3}{c|}{Task} & OneRestore~\cite{RN34} & +CURE & \multicolumn{3}{c|}{Task} & OneRestore & +CURE & \multicolumn{3}{c|}{Task} & OneRestore & +CURE \\
            \hline \hline
            Low$+$Haze  & Delow  & $\rightarrow$ & Dehaze & 20.95 & 25.80 & Dehaze & $\rightarrow$ & Delow  & 21.20 & 24.76 & Dehaze & $+$ & Delow  & 25.07 & 25.28 \\
            Low$+$Rain  & Delow  & $\rightarrow$ & Derain & 22.66 & 25.77 & Derain & $\rightarrow$ & Delow  & 24.45 & 25.54 & Derain & $+$ & Delow  & 25.27 & 25.41 \\
            Low$+$Snow  & Delow  & $\rightarrow$ & Desnow & 21.63 & 25.26 & Desnow & $\rightarrow$ & Delow  & 22.63 & 24.87 & Desnow & $+$ & Delow  & 24.64 & 24.95 \\
            Haze$+$Rain & Dehaze & $\rightarrow$ & Derain & 27.74 & 30.12 & Derain & $\rightarrow$ & Dehaze & 28.92 & 30.75 & Derain & $+$ & Dehaze & 29.62 & 30.52 \\
            Haze$+$Snow & Dehaze & $\rightarrow$ & Desnow & 26.26 & 28.66 & Desnow & $\rightarrow$ & Dehaze & 27.13 & 29.59 & Desnow & $+$ & Dehaze & 28.65 & 29.42 \\
            \cline{1-16}
            Average     & \multicolumn{3}{c|}{-} & 23.85 & \textbf{27.12} & \multicolumn{3}{c|}{-} & 24.87 & \textbf{27.10} & \multicolumn{3}{c|}{-} & 26.65 & \textbf{27.12} \\
            \hline
        \end{tabular}
    }
\end{table}
%

\subsection{Experiment Settings}

\label{sec:implementation_details}
We conduct the training on 4 NVIDIA A100 GPUs, requiring approximately two days for our method.
We initialize the process using the checkpoints of each baseline model where performance has reached saturation.
During the training of CURE, we set the learning rate to a value moderately lower than the baseline's initial setting.
This ensures that the rate is sufficient for learning new tasks yet conservative enough to maintain stability.

\noindent \textbf{Evaluation Dataset and Metrics.}
%
We create Controllable Composite Degradation Dataset (CCDD-11), which follows the pipeline introduced in CDD-11~\cite{RN34} (see Supp. Material for details).
Specifically, our CCDD-11 adopts the same pipeline as CDD-11, while generating all pairs corresponding to individual degradation removal from composite degradation images.
However, due to the limited variety of rain masks in the original CDD-11, we incorporate a more diverse range of rain masks into our CCDD-11.
%
%
Restoration performance is evaluated using PSNR and SSIM two standard metrics.

\noindent \textbf{Compared Methods.}
We compare our CURE with seven one-to-one models: MIRNet~\cite{zamir2020learning}, MPRNet~\cite{RN8}, MIRNetV2~\cite{zamir2022learning}, Restormer~\cite{RN25}, DGUNet~\cite{mou2022deep}, NAFNet~\cite{chen2022simple}, and Fourmer~\cite{zhou2023fourmer}; five one-to-many models: AirNet~\cite{RN29}, AdaIR~\cite{cui2025adair}, PromptIR~\cite{RN9}, WGWS~\cite{RN27}, and DFPIR~\cite{tian2025degradation}; and three Text-guided models: TextPromptIR~\cite{yan2025textual}, AM-PromptIR~\cite{zhang2025adaptive}, and OneRestore~\cite{RN34}.
All models are trained and evaluated on the CCDD-11 training and test sets.
%
During CURE training, we exclude images with triple degradations due to combinatorial complexity; however, these are included in the test set to assess generalization.
For PromptIR, as the correspondence between the number of learnable prompts and degradation types is unspecified, we evaluated both five and eleven prompts.
The version with five prompts achieves better performance and is thus reported.
For WGWS, weather-specific parameters are adjusted for CCDD-11, where WG and WS denote the first and second stages, respectively.
For AM-PromptIR and TextPromptIR, we align their input prompts with the degradation types in CCDD-11.
We construct these text prompts following the protocols specified in their respective papers.
\subsection{Comparisons}
As reported in~\Cref{tb:comparison}, the integration of our CURE framework consistently enhances restoration performance across all three models.
Beyond achieving superior quantitative metrics, our framework empowers existing architectures with fine-grained controllability.
%
%
Interestingly, although not originally designed as a One-to-Composite model, PromptIR outperforms OneRestore on the CCDD-11 dataset.
However, it is worth noting that PromptIR employs a much larger model and, by design, is not easily adaptable for controllability.
As shown in~\Cref{fig:comparison}, our CURE more effectively removes degradations compared to existing approaches in terms of visual quality.
Especially, in the second row of \Cref{fig:comparison}, the red box highlights a challenging case where both PromptIR and OneRestore leave noticeable snow artifacts, whereas our CURE achieves more effective removal under composite degradation ($low + haze + snow$).
We believe our disentangled learning approach provides more effective guidance, offering not only enhanced controllability but also superior performance in composite degradation restoration.

\subsection{Ablation Studies}
%
%
To verify the individual contributions of each component, we present an ablation analysis focusing on the proposed loss functions, identity embedding, and ratio-control embedding.
%
%
For detailed experimental settings and extra qualitative outputs, readers are directed to the accompanying Supplementary Material.

\noindent \textbf{Effect of Losses.}
To verify the effectiveness of the proposed loss functions, we conduct an ablation study by excluding each component, as shown in \Cref{tab:losses_ablation}.
While identity loss alone leads to improved performance compared to OneRestore~\cite{RN34}, incorporating additional loss functions results in further performance gains.
The best performance is achieved when all loss components are used together, indicating that the proposed loss functions effectively enable the text-guided restoration model to disentangle mixed degradations.

\noindent \textbf{Effect of Identity Embedding.}
The identity embedding is designed to enforce the preservation of the input image, even in the presence of degradation.
To verify its effectiveness, we measure the PSNR between the input and the output generated using the identity embedding.
For comparison, we evaluate OneRestore~\cite{RN34} with the "$clear$" prompt, which most closely matches the identity condition in its framework.
As shown in \Cref{tab:identity_psnr}, our CURE yields outputs with minimal deviation from the inputs, demonstrating superior identity preservation compared to OneRestore and validating our approach.

\noindent \textbf{Effect of Ratio-Control Embedding.}
The ratio-control embedding is designed to guide the restoration intensity by removing only a specified proportion of the degradation.
To achieve this, we construct a prompt by linearly combining the identity and target degradation embeddings according to~\Cref{eq:intensity_control_embedding}.
To quantitatively evaluate this controllability, we measure classification accuracy across varying values of $w$ in~\Cref{eq:intensity_control_embedding}.
Note that $w = 1$ leads to full degradation removal and low classifier accuracy, while $w = 0$ preserves the degradation and yields high accuracy.
As reported in \Cref{tab:intensity-ratio-accuracy}, the classification performance varies consistently with $w$, indicating that CURE effectively enables fine-grained intensity control.
This controllability is visually confirmed in \Cref{fig:qualitative-results}(a), which demonstrates the precise adjustment of restoration intensity.
\noindent \textbf{Selective Restoration.}
By enforcing explicit constraints on degradation independence through our disentangled learning objectives, our framework precisely isolates and removes a target degradation while preserving others.
This capability is crucial for addressing composite degradations.
We compare our method against OneRestore~\cite{RN34}, a strong baseline capable of handling composite degradations.
For instance, given an input with composite degradations of $low$ and $haze$, if the task is $delow$, the goal is to remove only the $low$-light effect while retaining the $haze$.
In \Cref{tab:partial_two_composite_psnr}, we report quantitative results on selective restoration tasks involving two degradations.
Here, the GT is synthesized by applying only the remaining degradation types to the clean image.
While the baseline supports single-degradation embeddings, it lacks explicit disentanglement supervision.
As shown in \Cref{tab:partial_two_composite_psnr}, our model achieves superior performance across all selective tasks.
We also validate our method on triple-composite scenarios.
Detailed results in the Supplementary Material confirm that our model maintains its performance advantage over the baseline even in these composite settings.
To further validate CURE's disentanglement capability, we conduct a classification-based analysis.
Specifically, we perform selective restoration on composite images to remove a target degradation and then classify the remaining degradation types in the output using a pre-trained classifier.
This protocol objectively verifies if non-target degradations are successfully preserved without distortion.
As shown in \Cref{tab:classification}, our method consistently outperforms the baseline in identifying the remaining degradations, regardless of the type being removed.
Moreover, qualitative comparisons in \Cref{fig:qualitative-results}(b) confirm the selective restoration closely matches the corresponding GT.
These results demonstrate CURE effectively decomposes composite degradations into their individual components for fine-grained control.

\noindent \textbf{Restoration-Order Dependency.}
For composite degradations, restoration can be performed by sequentially removing each degradation type.
As shown in \Cref{tab:order-dependent}, our method maintains consistent and high performance regardless of the restoration sequence, whereas OneRestore~\cite{RN34} exhibits significant performance variance depending on the order.
We attribute this order-invariant robustness to the proposed permutation-invariant loss defined in \Cref{eq:permutation-invariant}, which encourages the model to treat each degradation independently.
Notably, even when performing simultaneous restoration in a single step (using composite embeddings), our CURE outperforms the baseline, further validating its superior controllability.
This can also be visually confirmed in \Cref{fig:qualitative-results}(c).

\noindent \textbf{Blur-Noise-JPEG Degradation.}
To validate generalizability beyond weather, we evaluate CURE on digital degradations ($blur$, $noise$, and $JPEG$)
It is worth noting that CURE consistently outperforms the baselines in this setting as well.
As illustrated in \Cref{fig:qualitative-results}(d), our method achieves successful restoration by effectively removing complex digital artifacts while preserving image details.
We also perform the same evaluations as CCDD-11 including intensity analysis, classification accuracy, selective restoration, and restoration order to ensure consistency. 
{Furthermore, we explore the model's behavior under zero-shot and unseen degradations by testing it on real-world datasets. }
Detailed quantitative results and real-world evaluations are provided in the Supplementary Material.

\section{Conclusion}
\label{sec:conclusion}
In this paper, we propose CURE, a disentangled learning framework for controllable all-in-one image restoration under composite degradation scenarios. 
By integrating four specialized objectives, CURE enables users to flexibly select both the type and degree of restoration without altering unrelated image content. 
Our proposed method supports fine-grained user-guided intensity control and achieves highly competitive performance on composite degradation benchmarks.
However, a remaining limitation is that while it is effective for up to three concurrent degradations, performance may drop when scaling to a larger number of types, particularly with complex spatially varying degradations.
Addressing this challenge will further enhance its scalability in real-world applications.

%
%

\section*{Acknowledgements}
This work was partly supported by Institute of Information \& communications Technology Planning \& Evaluation (IITP) grant funded by the Korea government(MSIT) (No.RS-2025-25422680, Metacognitive AGI Framework and its Applications, 25 \%), Institute of Information \& communications Technology Planning \& Evaluation (IITP) grant funded by the Korea government(MSIT) (No.RS-2020-II201373, Artificial Intelligence Graduate School Program(Hanyang University), 25 \%)
and Korea Institute for Advancement of Technology(KIAT) grant funded by the Korea Government(MOTIE) (RS-2024-00415938, HRD Program for Industrial Innovation 25 \%) and the National Research Foundation of Korea(NRF) grant funded by the Korea government(MSIT) (RS-2025-00521432, 25 \%).

\title{Supplementary Material of "CURE: Controllable Unified Image Restoration for Complex Degradations"}
\titlerunning{CURE: Controllable Unified Image Restoration for Complex Degradations}

\author{Boseong Kim \and Donghyeon Cho\thanks{Corresponding author.}}
\institute{Department of Computer Science, Hanyang University, Seoul, South Korea \\
\email{\{hosky789, doncho\}@hanyang.ac.kr}}

\maketitle
%
\section{Controllable Composite Degradation Dataset}
\label{sec:ccdd}
\subsection{Dataset Construction Necessity}
Our work addresses controllable restoration tasks that existing datasets cannot support. 
Specifically, CURE requires ground-truth pairs that are absent in existing public datasets: (1) half-intensity degradation images with identical content and context for ratio control training, and (2) selective restoration pairs where specific degradations are removed from composite images while preserving others.
Without these pairs, it is impossible to train or evaluate models on fine-grained controllability tasks.
Therefore, the Controllable Composite Degradation Dataset (CCDD-11) is created not as an optional addition, but as an essential foundation for controllable composite image restoration research.
To ensure objectivity, CCDD-11 strictly follows the original CDD-11 synthesis methodology with transparent, reproducible extensions using fixed random seeds and uniform parameter distributions.

\subsection{CCDD-11 Data Pipeline}
To construct CCDD-11, we generally follow the protocol of the CDD-11~\cite{RN34} dataset, with slight modifications to enable controllable restoration. 
Specifically, we introduce selective restoration pairs and half-restoration pairs, which allow for more flexible and targeted restoration scenarios.
To further enrich the dataset, we incorporate a wider variety of rain masks, which increases the diversity of degradation patterns in CCDD-11.
For the source images, we select 1,383 high-resolution clean images from the RAISE~\cite{dang2015raise} dataset and uniformly resize them to a resolution of 1080×720 pixels.
Of these, 1,183 images are used for training and the remaining 200 for testing.
The procedure for generating composite degradation images is as follows:
\begin{equation}
    I(x) = \mathcal{D}_h(\mathcal{D}_{rs}(\mathcal{D}_l(J(x)))), \quad
    \label{eq:compsite-data}
\end{equation}
where $I(x)$ denotes the degraded image, $J(x)$ is the corresponding clean image as described above, $\mathcal{D}_h$ represents haze degradation, $\mathcal{D}_{rs}$ represents rain or snow degradation, and $\mathcal{D}_l$ represents low-light degradation.
Each of $\mathcal{D}_h$, $\mathcal{D}_{rs}$, and $\mathcal{D}_l$ can be applied independently or omitted during the degradation process.
The order and combination of these degradations follow the protocol established in CDD-11.
Note that in addition to the fully composite degraded images generated by Eq.~\eqref{eq:compsite-data}, we also store images degraded by only a subset of these degradations.
For these subset images, we use exactly the same degradation masks and parameters as in the composite degradations, ensuring that they can serve as ground truth for selective restoration tasks.
For example, when generating an image with $haze+rain$ degradations, we also store the corresponding $haze$ and $rain$ images, which are produced using the same degradation masks and parameters, following the same degradation pipeline.
We also define half-degradation images as those generated by applying the same degradation process but with approximately half the intensity of the original parameters. 
These half-degradation images are also stored alongside the fully degraded images to enable our proposed intensity ratio control restoration.

\noindent \textbf{Low-Light.}
According to Retinex theory, the low-light image generation pipeline is defined as follows:
\begin{equation}
    I_{l}(x) = \mathcal{D}_l(x) =\frac{J(x)}{L(x)} \cdot {L(x)}^\gamma + \varepsilon,
    \label{eq:low}
\end{equation}
%
where $J(x)$ denotes the clean image and $I_l(x)$ is the corresponding low-light image.
In this formulation, $\gamma$ is a darkening coefficient ranging from 2 to 3, which serves as a brightness adjustment factor for the illumination map $L(x)$ generated by LIME~\cite{guo2016lime}.
Gaussian noise $\varepsilon$ with zero mean and variance between 0.03 and 0.08 is added to better simulate low-light environments.
%
\begin{equation}
    I_{l_h}(x) = \frac{J(x)}{L(x)} \cdot {L(x)}^{\gamma/2} + \varepsilon,
    \label{eq:half-low}
\end{equation}
%
where $I_{l_h}(x)$ denotes the half-low-light image, which is generated by applying the low-light pipeline with the $\gamma$ parameter set to half the value used in Eq.~\eqref{eq:low}.

\noindent \textbf{Rain/Snow Streaks.}
Following~\cite{chen2021all}, we add snow streaks to images using alpha blending.
Similarly, our rain synthesis, based on~\cite{li2019heavy}, is modified to use alpha blending, enabling ratio control through adjustable weighting.
The pipeline for synthesizing rain and snow streaks is as follows:
\begin{equation}
    I_{rs}(x) = D_{rs}({D}_l(x)) = {I}_l(x)(1-\mathcal{RS}) + \mathcal{RS},
    \label{eq:rain-snow}
\end{equation}
%
where $I_{rs}(x)$ denotes the rain/snow streak image, and $\mathcal{RS}$ represents rain mask or snow mask.
The rain mask is sourced from~\cite{garg2006photorealistic}, while the snow mask is sourced from~\cite{liu2018desnownet}.
%
\begin{equation}
    I_{rs_h}(x) = {I}_l(x)(1-0.5 \cdot \mathcal{RS}) + 0.5 \cdot \mathcal{RS},
    \label{eq:half-rain-snow}
\end{equation}
%
where $I_{rs_h}(x)$ denotes the half-intensity rain/snow streak image, generated by applying alpha blending with the weight for rain or snow reduced to half of that used in Eq.~\eqref{eq:rain-snow}.

\noindent \textbf{Haze.}
Haze degradation is introduced into our pipeline using the atmospheric scattering model, as follows:
%
\begin{equation}
    t = e^{-\beta \cdot d(x)}, \quad
    t_{h} = e^{-(\beta/2) \cdot d(x)},
    \label{eq:transmission1}
\end{equation}
where $t$ is the transmission map, $\beta$ is the haze density coefficient, and $d(x)$ is the depth information estimated from MegaDepth~\cite{li2018megadepth}.
Here, $t_{h}$ represents the half-transmission map, which is obtained by setting the haze density coefficient to $\beta/2$.
The value of $\beta$, which controls the haze density, is set in the range [1.0, 2.0].
%
\begin{equation}
    I_{h}(x) = D_{h}(D_{rs}({D}_l(x))) = I_{rs}(x)t+A(1-t), 
    \label{eq:haze_pipeline}
\end{equation}
%
where $I_{h}(x)$ is the haze-degraded image, and $A$ is the atmospheric light, which is constrained to the range [0.6, 0.9].
%
\begin{equation}
    I_{h_h}(x) = I_{rs}(x)t_{h}+A(1-t_{h}), \quad
    \label{eq:haze}
\end{equation}
%
where $I_{h_h}(x)$ is the half-haze image, in which the haze density is approximately half that of $I_{h}(x)$ due to the use of $t_{h}$.
\subsubsection{Generalization and Diversity Analysis.}
To evaluate generalization capabilities of CCDD-11 compared to the original CDD-11, we conduct comprehensive cross-evaluation experiments using degradation classifiers (the Text/Visual Embedder of~\cite{RN34}). 
During our analysis, we observe that the original CDD-11 relies on a limited and repetitive set of rain masks. 
In contrast, CCDD-11 addresses this limitation by incorporating 4,723 unique rain masks, significantly increasing the complexity of the degradation patterns. 
This fundamental difference in dataset construction is clearly reflected in the cross-evaluation results shown in Tab.~\ref{tb:rain_relateddegradation_classification}. 
Classifiers trained on CCDD-11 maintain robust performance when evaluated on CDD-11, with the average accuracy dropping only slightly from 92\% to 88\%. 
In contrast, the same architecture trained on CDD-11 shows a severe performance collapse when tested on the more diverse CCDD-11, with accuracy plummeting from 98\% to 59\%.
This asymmetric generalization pattern demonstrates that models trained on the restricted patterns of CDD-11 fail to handle diverse rain scenarios, whereas CCDD-11 provides a superior training signal that yields models robust to both simple and complex degradations.
\begin{table}[t]
    \centering
    \small
    \caption{Classification accuracy of rain-related degradation under cross-dataset evaluation between CCDD-11 and CDD-11.}
    \label{tb:rain_relateddegradation_classification}
    \resizebox{\linewidth}{!}{
        \begin{tabular}{l|c|c|c|c}
            \hline
            \textbf{Degradation Type} & \textbf{(1) Trained on CCDD-11} & \textbf{(2) Trained on CCDD-11} & \textbf{(3) Trained on CDD-11} & \textbf{(4) Trained on CDD-11} \\
            & \textbf{Test on CCDD-11} & \textbf{Test on CDD-11} & \textbf{Test on CCDD-11} & \textbf{Test on CDD-11} \\
            \hline \hline
            rain & 98 \% & 96 \% & 75 \% & 99 \% \\
            low rain & 99 \% & 86 \% & 58 \% & 97 \% \\
            haze rain & 95 \% & 90 \% & 58 \% & 98 \% \\
            low haze rain & 76 \% & 82 \% & 46 \% & 95 \% \\
            \hline
            \textbf{Average} & \textbf{92 \%} & \textbf{88 \%} & \textbf{59 \%} & \textbf{98 \%} \\
            \hline
        \end{tabular}
    }
\end{table}
\subsection{Blur-Noise-JPEG Dataset}
To demonstrate the generalizability of our proposed method, we construct not only the weather-related CCDD-11 dataset described above but also an additional dataset incorporating digital degradations, such as blur, noise, and JPEG compression.
In constructing the Blur-Noise-JPEG dataset, degradations are applied in the order of blur, noise, and JPEG compression.
This sequence reflects the typical image formation process, in which optical blur occurs first, followed by sensor noise and then compression, as described in recent works such as Real-ESRGAN~\cite{wang2021real}.
\begin{equation}
    I(x) = \mathcal{D}_J(\mathcal{D}_{n}(\mathcal{D}_b(J(x)))),
    \label{eq:bnj-data}
\end{equation}
where $I(x)$ denotes the final degraded image, $J(x)$ is the corresponding clean image, $\mathcal{D}_b$ represents blur degradation, $\mathcal{D}_n$ denotes the addition of Gaussian noise, and $\mathcal{D}_J$ indicates JPEG compression.
The degradations are applied sequentially in the order of blur, noise, and JPEG compression, resulting in seven types of degradation in the dataset: Blur, Noise, JPEG, Blur$+$Noise, Blur$+$JPEG, Noise$+$JPEG, and Blur$+$Noise$+$JPEG.
Similar to CCDD-11, the Blur-Noise-JPEG dataset also stores both subset degradation images and half-degradation images for each sample.

\noindent \textbf{Blur.}  
Blur degradation simulates the loss of sharpness that typically results from camera defocus or motion.
The process for generating blurred images is defined as follows:
\begin{equation}
    I_{b}(x) = J(x) * G(x; k, \sigma_b), \qquad
    I_{b_h}(x) = J(x) * G(x; k, \sigma_b/2),
    \label{eq:blur}
\end{equation}
where $I_b(x)$ and $I_{b_h}(x)$ denote blurred and half-blurred images respectively, $J(x)$ is the clean image, $*$ represents the convolution operation, and $G(x; k, \sigma_b)$ is a 2D Gaussian kernel with kernel size $k$ and standard deviation $\sigma_b$. 
The half-blurred image is generated using a standard deviation of $\sigma_b/2$.

\noindent \textbf{Noise.}  
Noise degradation simulates random pixel fluctuations resulting from sensor imperfections during image acquisition, and is defined as follows:
\begin{equation}
    I_{n}(x) = J(x) + \mathcal{N}(0, \sigma_n^2), \qquad
    I_{n_h}(x) = J(x) + \mathcal{N}(0, (\sigma_n/2)^2),
    \label{eq:noise}
\end{equation}
where $I_n(x)$ and $I_{n_h}(x)$ denote noisy and half-noisy images, respectively.
In this equation, $J(x)$ is the clean image, and $\mathcal{N}(0, \sigma_n^2)$ denotes zero-mean Gaussian noise with standard deviation $\sigma_n$.
The half-noisy image is generated using a standard deviation of $\sigma_n/2$.

\noindent \textbf{JPEG.}  
JPEG degradation simulates compression artifacts that are typically introduced during image encoding and storage.
The process for generating JPEG-compressed images is defined as follows:
\begin{equation}
    I_{j}(x) = \mathrm{JPEG}(J(x);\, q), \qquad
    I_{j_h}(x) = \mathrm{JPEG}(J(x);\, q/2),
    \label{eq:jpeg}
\end{equation}
where $I_j(x)$ and $I_{j_h}(x)$ denote the JPEG-compressed and half-JPEG images, respectively. In this equation, $J(x)$ is the clean image, and $\mathrm{JPEG}(J(x); q)$ denotes the operation of compressing $J(x)$ using the JPEG algorithm with quality factor $q$.
The half-JPEG image is generated using a quality factor of $q/2$.
%

%
%
\section{Training Details}
\label{sec:training-details}
%

\noindent \textbf{Baseline Saturation.}
We adopt a two-stage training strategy to ensure that the model first establishes robust restoration capabilities before learning fine-grained intensity control. 
In the first stage, we train each baseline model including AM-PromptIR~\cite{zhang2025adaptive}, TextPromptIR~\cite{yan2025textual}, and OneRestore~\cite{RN34} by following their respective original implementation. 
This includes using their original optimizers, initial learning rates, and schedulers. 
Each baseline is trained until its performance on the composite image restoration task reaches saturation.
For instance, we train OneRestore for up to 300 epochs to ensure the base architecture is fully optimized for handling complex degradations before we incorporate our CURE.

\noindent \textbf{Integration of CURE.}
Once the baseline performance is saturated, we resume training from the best performing checkpoint to incorporate our proposed CURE. 
We carefully calibrate the learning rate for this stage to ensure stable adaptation. 
Specifically, the learning rate is set lower than the initial learning rate of the baseline but higher than the final decayed rate at the point of saturation.
This policy allows the model to learn the disentanglement and control mechanisms without forgetting the previously learned restoration capabilities.

\noindent \textbf{Implementation details.}
Throughout the training process, all architectural hyperparameters and loss weights from the original baselines are kept unchanged to ensure a fair comparison.
For example, the weight ratios of $0.6, 0.3, 0.1$ for OneRestore are maintained throughout.
For the CCDD-11 dataset, we use 1,183 images with 11 different degradation types, which results in a total of 13,013 training samples.
The models are evaluated on a separate test set of 200 images per degradation type.
All experiments are implemented in PyTorch and trained until peak performance is achieved. 
This typically requires an additional 60 to 200 epochs depending on the complexity of the baseline model.
%

\section{Experimental Details}
\label{sec:experimental-detail}
This section provides a detailed description of the experiments on the CCDD-11 dataset, extending the discussion presented in the main paper.
To evaluate the intensity ratio control performance (Table 4 in the main paper), we follow the experimental procedure described below.
For each degradation type, we vary the ratio control embedding from 0.0 to 1.0 and perform restoration accordingly.
Each restored result is then classified using the Text/Visual Embedder, which can classify the remaining degradation types. 
The reported value is the average classification accuracy across all degradation types and intensity ratios.
This setup is motivated by the fact that when $w=0.0$ (i.e., identity), the image remains unchanged from the input degradation, and thus is expected to be classified with high accuracy as the corresponding degradation type.
In contrast, when $w=1.0$, the model restores the image to be close to a clean image, which should result in low classification accuracy.
This approach is adopted because generating ground truth pairs for every possible value of $w$ is impractical.
The experiments in Tab.~\ref{tab:partial_two_composite_psnr_bnj} and Tab.~\ref{tab:partial_three_composite_psnr_bnj} are designed to demonstrate that our CURE can perform selective restoration, accurately removing only the targeted degradations while preserving the others.
In Tab.~\ref{tab:partial_two_composite_psnr_bnj}, we evaluate the PSNR performance of restoring each individual degradation from composite images containing two types of degradation, using ground-truth pairs generated according to our dataset pipeline.
Tab.~\ref{tab:partial_three_composite_psnr_bnj} extends this experiment to composite images containing three types of degradation.
In this setting, we consider two tasks: double restoration, which targets restoring two out of the three degradations (leaving one degradation in the image), and single restoration, which targets restoring only one degradation (leaving the other two degradations in the image).
Since there are many possible combinations for each task, we report the average PSNR across all cases.
The experiment in Tab.~\ref{tab:order-dependent_bnj} is designed to demonstrate the mitigation of order dependency by comparing the performance of two-stage and one-stage restoration on composite images containing two types of degradation.
For the composite degradation, Blur$+$Noise, the notations Deblur $\rightarrow$ Denoise and Denoise $\rightarrow$ Deblur represent two-stage restoration, where each degradation is removed sequentially in a different order.
In contrast, Denoise $+$ Deblur refers to the one-stage restoration, where both degradations are removed simultaneously in a single step.
%
\section{Additional Experimental Results}
\subsection{Extended Results on CCDD-11}
As discussed in the main paper, we extend our evaluation of selective restoration to more complex triple-composite scenarios. 
Tab.~\ref{tab:triple_partial_psnr} presents the quantitative results on the CCDD-11 dataset involving three simultaneous degradations ($Low+Haze+Rain/Snow$).
Notably, despite not being explicitly trained on triple-composite scenarios, CURE achieves a substantial performance lead in both Double and Single Restoration tasks (restoring two and one components, respectively).
Thus, CURE generalizes to higher-order complexity without direct supervision, providing a robust training signal that isolates target components.
%
\begin{table}[t]
    \centering
    \caption{PSNR evaluation of selective restoration on the CCDD-11 dataset with triple composite degradations.}
    \label{tab:triple_partial_psnr}
    \resizebox{0.85\linewidth}{!}{
        \begin{tabular}{c|c|c|c|c}
            \hline
            \multicolumn{1}{c|}{Degradation}  & \multicolumn{2}{c|}{Double Composite Restoration} & \multicolumn{2}{c}{Single Restoration} \\
            \cline{2-5}
             & OneRestore~\cite{RN34} & +CURE & OneRestore & +CURE \\
            \hline \hline
            Low$+$Haze$+$Rain & 22.33 & 27.00 & 25.10 & 32.50 \\
            Low$+$Haze$+$Snow & 21.29 & 25.42 & 24.27 & 31.48\\
            \cline{1-5}
            Average & 21.81 & \textbf{26.21} & 24.69 & \textbf{31.99}\\
            \hline
        \end{tabular}
        \label{tab:partial_three_composite_psnr}
    }
\end{table}
%
\subsection{Evaluation on the Blur-Noise-JPEG Dataset}
\label{sec:blur-noise-jpeg}
%
\begin{table*}[t]
    \centering
    \footnotesize
    \caption{Quantitative comparisons on Blur-Noise-JPEG dataset.}
    \label{tab:comparison_blur_noise_jpeg}
        \begin{tabular}{l|c|c|c|c}
            \hline
            Types & Methods & PSNR $\uparrow$ & SSIM $\uparrow$ & \#Params \\
            \hline \hline
            & Input & 25.82 & 0.66 & - \\
            \hline \hline
            Text-guided
            & AM-PromptIR~\cite{zhang2025adaptive} & 28.62 & 0.90 & 64.54M \\ 
            & TextPromptIR~\cite{yan2025textual} & 28.91 & 0.83 & 93.55M \\ 
            & OneRestore (Visual)~\cite{RN34} & 27.53 & 0.79 & 5.98M \\ 
            & OneRestore (Text)~\cite{RN34} & 27.59 & 0.80 & 5.98M \\
            \hline
            Fine Control
            & AM-PromptIR$+$ CURE & {28.81} & 0.89 & 64.54M \\ 
            & TextPromptIR $+$ CURE & {28.97} & 0.83 & 93.55M \\ 
            & OneRestore $+$ CURE (Visual) & {28.18} & 0.81 & 5.98M \\ 
            & OneRestore $+$ CURE (Text) & {28.21} & 0.81 & 5.98M \\
            \hline
        \end{tabular}%
    
\end{table*}
\begin{table}[t]
        \centering
        \caption{Embedder classification accuracy according to changes in the intensity ratio on Blur-nois-JPEG dataset.}
        \label{tab:intensity-ratio-accuracy_bnj}
        \resizebox{0.95\linewidth}{!}{
        \begin{tabular}{c|c|c|c|c|c|c|c|c|c|c|c}
            \hline
            Intensity Ratio $w$ & 1.0 & 0.9 & 0.8 & 0.7 & 0.6 & 0.5 & 0.4 & 0.3 & 0.2 & 0.1 & 0.0 \\
            \hline \hline
            Accuracy & 1.5 \% & 1.9 \% & 4.1 \% & 11.7 \% & 16.4 \% & 17.7 \% & 18.8 \% & 36.0 \% & 54.4 \% & 64.8 \% & 63.8 \% \\
            \hline
        \end{tabular}}
\end{table}
%
\begin{table}[t]
    \centering
    \caption{PSNR evaluation of selective restoration on images with two combined degradations.}
    \label{tab:partial_two_composite_psnr_bnj}
    \resizebox{0.9\linewidth}{!}{
    \begin{tabular}{c||c|c|c||c|c|c}
        \hline
        Degradation & Task & OneRestore~\cite{RN34} & +CURE & Task & OneRestore~\cite{RN34} & +CURE \\
        \hline \hline
        Blur$+$Noise & Deblur & 24.87 & 28.56 & Denoise & 36.21 & 37.99\\
         Blur$+$JPEG &  Deblur & 24.71 & 26.92 & DeJPEG & 37.89 & 39.98 \\
         Noise$+$JPEG &  Denoise & 27.77 & 30.32 & DeJPEG & 24.74 & 25.02\\
        \cline{1-7}
        Average & - & 25.78 & \textbf{28.60} & - & 32.95 & \textbf{34.33} \\
        \hline
        \hline
    \end{tabular}
    }
\end{table}
%
\begin{table}[t]
    \centering
    \caption{PSNR evaluation of selective restoration on Blur-Noise-JPEG dataset triple composite degradations.}
    \label{tab:partial_three_composite_psnr_bnj}
    \resizebox{0.8\linewidth}{!}{
    \begin{tabular}{c|c|c|c|c}
        \hline
        \multicolumn{1}{c|}{Degradation}  & \multicolumn{2}{c|}{Double Restoration} & \multicolumn{2}{c}{Single Restoration} \\
        \cline{2-5}
         & OneRestore~\cite{RN34} & +CURE & OneRestore~\cite{RN34} & +CURE \\
        \hline \hline{}
        Blur$+$Noise$+$JPEG & 27.22 & 28.09 & 27.36 & 29.65 \\
        \hline
        \hline
    \end{tabular}
    }
\end{table}

\begin{table}[t]
    \centering
        \centering
        \caption{PSNR evaluation with varying restoration orders.}
        \label{tab:order-dependent_bnj}
        \resizebox{0.99\linewidth}{!}{
        \begin{tabular}{c||c|c|c||c|c|c||c|c|c}
            \hline
            Degradation & Task & ~\cite{RN34} & Ours & Task & ~\cite{RN34} & Ours & Task & ~\cite{RN34} & Ours \\
            \hline \hline
            Blur$+$Noise & Deblur $\rightarrow$ Denoise & 21.97 & 26.01 & Denoise $\rightarrow$ Deblur & 23.95 & 25.99 & Deblur $+$ Denoise & 25.24 & 25.33\\
             Blur$+$JPEG &  Deblur $\rightarrow$ DeJPEG & 22.32 & 25.92 & DeJPEG $\rightarrow$ Deblur & 23.73 & 25.95 & Deblur $+$ DeJPEG & 25.07 & 25.14\\
             Noise$+$JPEG &  Denoise $\rightarrow$ DeJPEG & 27.02 & 29.41 & DeJPEG $\rightarrow$ Denoise & 26.94 & 29.42 & Denoise $+$ DeJPEG & 28.80 & 28.84\\
            \cline{1-10}
            Average & - & 23.77 & \textbf{27.11} & - & 24.87 & \textbf{27.12} & - & 26.37 & \textbf{26.44}  \\
            \hline
            \hline
        \end{tabular}}
\end{table}
To further verify the generalizability of CURE, we conduct extensive experiments on the Blur-Noise-JPEG dataset. Tab.~\ref{tab:comparison_blur_noise_jpeg} summarizes the quantitative comparisons.
Consistent with the results on CCDD-11 presented in the main paper, the integration of CURE leads to consistent performance gains across all text-guided baseline models.
These results demonstrate that CURE is a model-agnostic framework that effectively enhances restoration quality while providing fine-grained control capabilities.
In the subsequent detailed evaluations, we use OneRestore~\cite{RN34} as the representative baseline to analyze the specific functionalities of CURE.
Tab.~\ref{tab:intensity-ratio-accuracy_bnj} shows the ratio control accuracy on this dataset.
Although the absolute accuracy is lower than that on CCDD-11 due to the higher difficulty of the Blur-Noise-JPEG dataset, the accuracy still exhibits a clear linear trend as the parameter $w$ varies.
This result confirms that our mechanism effectively manipulates restoration intensity in this domain.
As shown in Tab.~\ref{tab:partial_two_composite_psnr_bnj} and Tab.~\ref{tab:partial_three_composite_psnr_bnj}, CURE also maintains superior performance in selective restoration tasks.
Additionally, the results in Tab.~\ref{tab:order-dependent_bnj} indicate that CURE effectively reduces performance variations caused by order dependency.
%
\section{Experimental Results on Real-world Datasets}
\label{sec:real-world}
\begin{figure}[t]
    \begin{center}
        \def\arraystretch{0.4}
        \begin{tabular}{@{}c}    
            \includegraphics[width=0.9\linewidth]{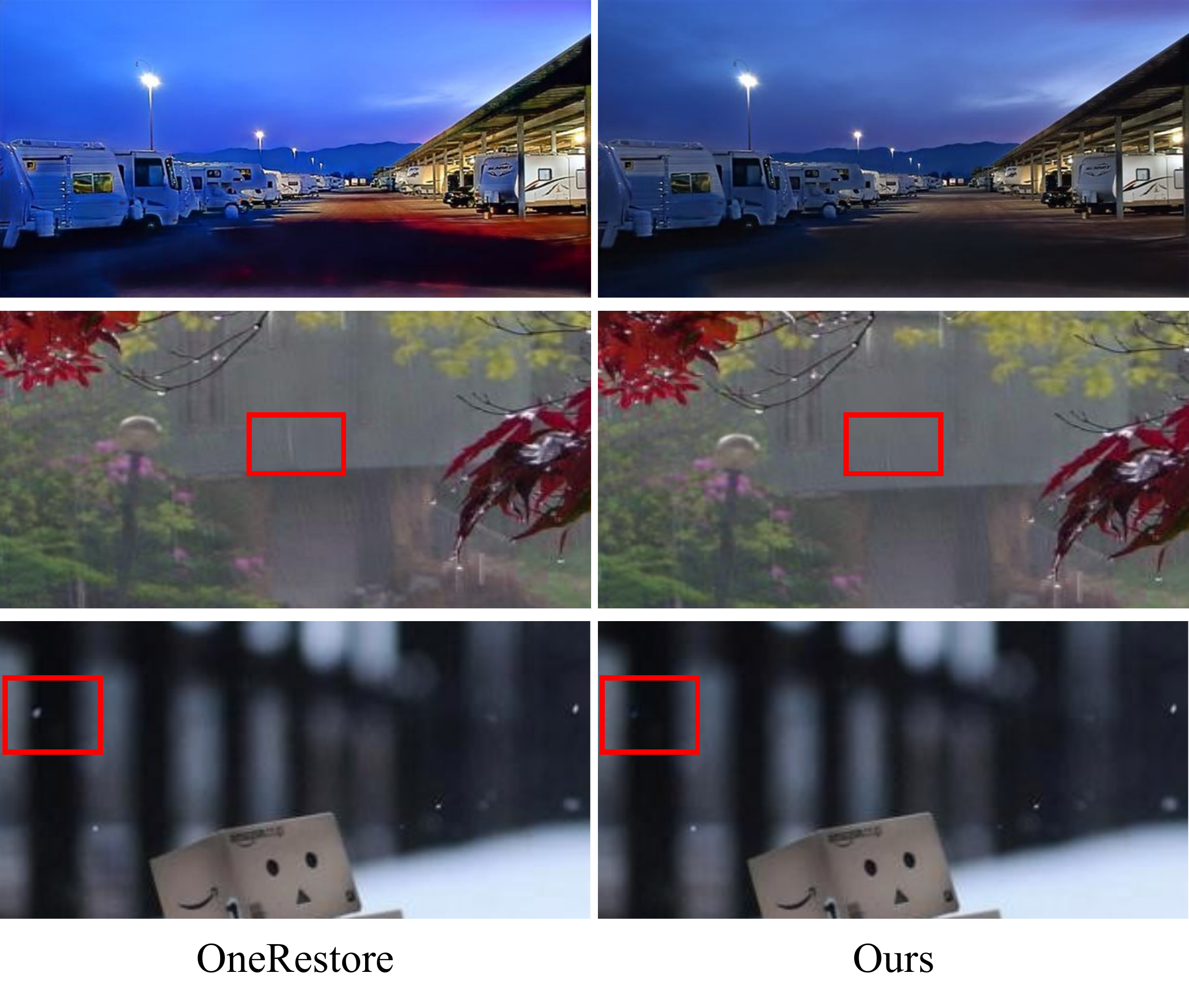}
        \end{tabular}
    \end{center}
    \caption{Qualitative Restoration Results on Real-World Datasets.}
    \label{fig:real}
\end{figure}

We assess the robustness of our model trained on CDD-11 by conducting extensive real-world image restoration experiments under challenging degradation scenarios.
To comprehensively address four distinct real-world degradations, we select the following benchmarks: NPE~\cite{wang2013naturalness} for low-light enhancement, RTTS~\cite{li2018benchmarking} for dehazing, RS~\cite{yang2017deep} for deraining, and Snow100k-R~\cite{liu2018desnownet} for desnowing.
Quantitative results are reported on these four real-world benchmarks using the no-reference quality metric Natural Image Quality Evaluator (NIQE).
Tab.~\ref{tab:real} shows that our model achieves competitive results in multiple datasets.
Specifically, our CURE outperforms previous approaches on RTTS and Snow100k-R, and achieves marginal improvements over OneRestore~\cite{RN34} on NPE and RS.
The qualitative results in Fig.~\ref{fig:real} also show that our CURE produces better restoration quality than the previous approach.
%
\begin{table}[t]
    \centering
    \caption{NIQE scores on four Real-world datasets.}
    \label{tab:real}
    \resizebox{0.99\linewidth}{!}{ 
        \begin{tabular}{c|c|c|c||c|c|c|c}
            \hline
            \multicolumn{2}{c|}{NPE~\cite{wang2013naturalness}} 
            & \multicolumn{2}{c||}{RS~\cite{yang2017deep}} 
            & \multicolumn{2}{c|}{Snow-100K-R~\cite{liu2018desnownet}} 
            & \multicolumn{2}{c}{RTTS~\cite{li2018benchmarking}} \\
            \cline{1-8}
            Method                & NIQE $\downarrow $
            & Method              & NIQE $\downarrow $
            & Method              & NIQE $\downarrow $
            & Method              & NIQE $\downarrow $ \\
            \hline \hline
            RUAS~\cite{liu2021retinex}       & 7.77 
            & MPRNet~\cite{zamir2021multi}   & 3.55 
            & DRT~\cite{liang2022drt} & 3.93 
            & MSBDN~\cite{dong2020multi} & 4.77 \\
            SCI~\cite{ma2022toward}       & \textbf{3.97}
            & DualGCN~\cite{fu2021rain}   & \textbf{3.27}
            & TUM~\cite{chen2022learning} & 3.13 
            & DeHamer~\cite{guo2022image} & 5.34 \\
            SNRNet~\cite{xu2022snr}       & 4.49 
            & MFDNet~\cite{wang2023multi} & 3.32
            & UMWT~\cite{kulkarni2022unified} & 3.34 
            & C2PNet~\cite{zheng2023curricular} & 5.03 \\
            OneRestore~\cite{RN34}        & 4.83 
            & OneRestore~\cite{RN34}      & 3.67 
            & OneRestore~\cite{RN34}      & 2.93 
            & OneRestore~\cite{RN34}      & 4.76 \\
            OneRestore+CURE               & 4.44    
            & OneRestore+CURE             & 3.66 
            & OneRestore+CURE            & \textbf{2.92}
            & OneRestore+CURE            & \textbf{4.53} \\
            \hline
            \hline
        \end{tabular}
    }
\end{table}
%
%
%
\section{Qualitative Evaluation Results}
\label{sec:qualitative-results}
\subsection{Identity Operation}
\begin{figure}[t]
    \begin{center}
        \def\arraystretch{0.4}
        \begin{tabular}{@{}c}    
            \includegraphics[width=0.9\linewidth]{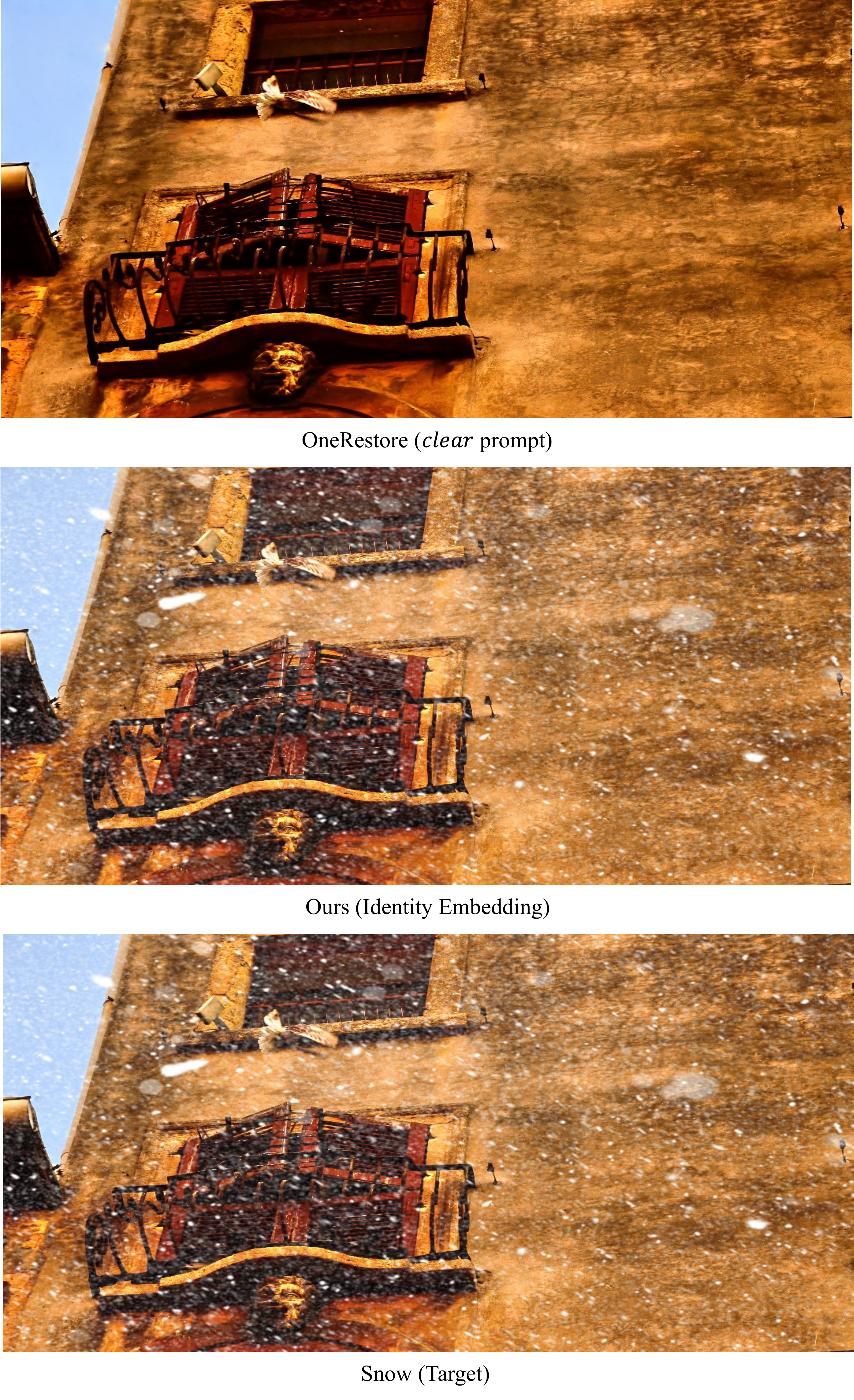}
        \end{tabular}
    \end{center}
    \caption{Qualitative comparison of identity operation for Snow image, where the Snow image serves as both the input and the target.}
    \label{fig:identity-1}
\end{figure}

\begin{figure}[t]
    \begin{center}
        \def\arraystretch{0.4}
        \begin{tabular}{@{}c}    
            \includegraphics[width=0.9\linewidth]{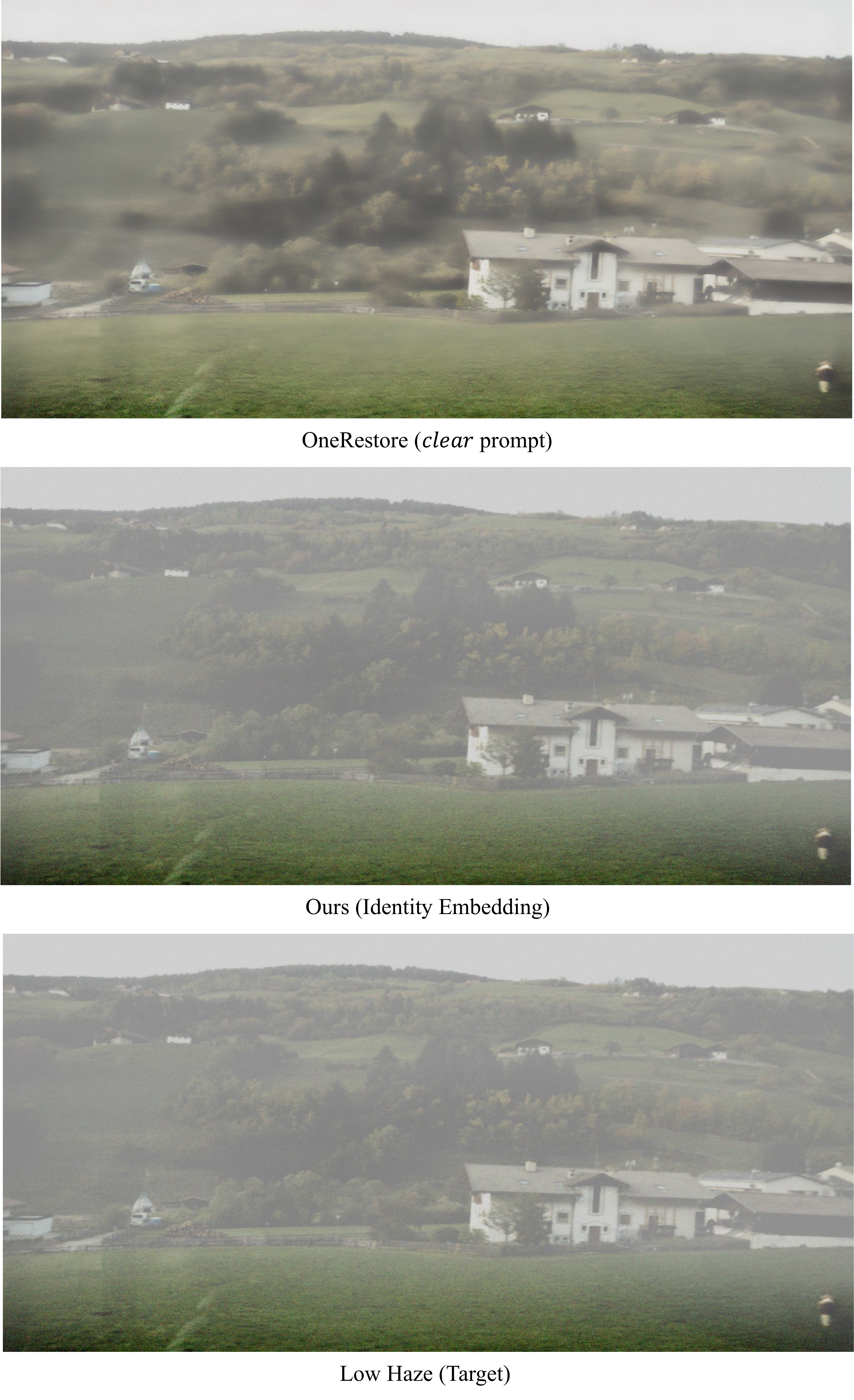}
        \end{tabular}
    \end{center}
    \caption{Qualitative comparison of identity operation for Low$+$Haze image, where the Low$+$Haze image serves as both the input and the target.}
    \label{fig:identity-2}
\end{figure}

To evaluate the identity operation, we present qualitative results showing that our model produces outputs identical to the inputs by employing the proposed identity embedding and identity loss.
As shown in the qualitative results in Fig.~\ref{fig:identity-1} and Fig.~\ref{fig:identity-2}, our CURE successfully preserves the input without introducing unwanted alterations, as intended.
For comparison, following the approach described in the main paper, we use the $clear$ prompt in OneRestore~\cite{RN34}, which serves as the closest equivalent to the identity condition in its framework.
In our method, we employ the proposed identity embedding.

\subsection{Ratio Control Restoration}
\begin{figure}[t]
    \begin{center}
        \def\arraystretch{0.4}
        \begin{tabular}{@{}c}    
            \includegraphics[width=0.9\linewidth]{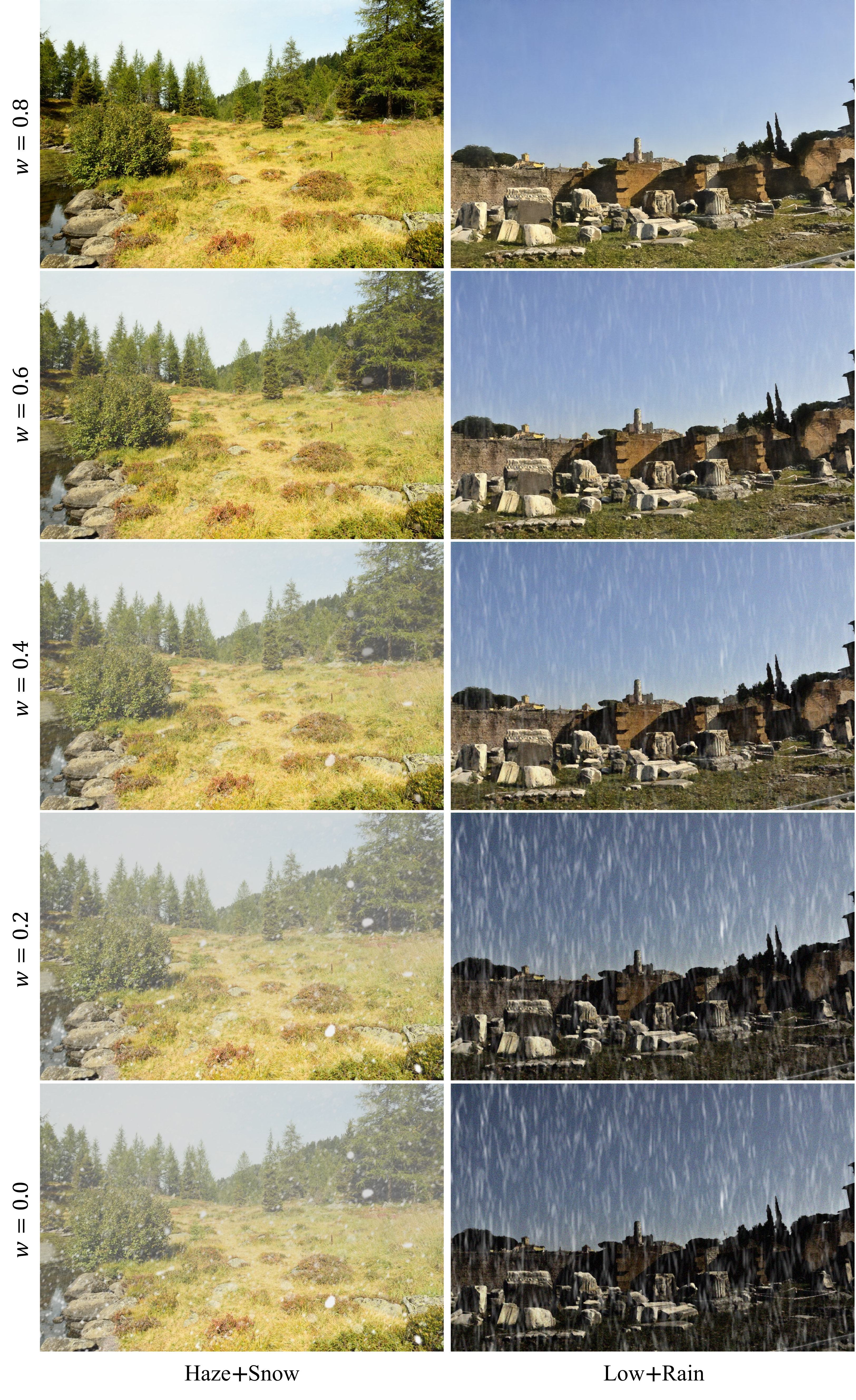}
        \end{tabular}
    \end{center}
    \caption{Qualitative results of Ratio Control Embedding with varying $w$ values for Haze$+$Snow and Low$+$Rain.}
    \label{fig:ratio-1}
\end{figure}

\begin{figure}[t]
    \begin{center}
        \def\arraystretch{0.4}
        \begin{tabular}{@{}c}    
            \includegraphics[width=0.9\linewidth]{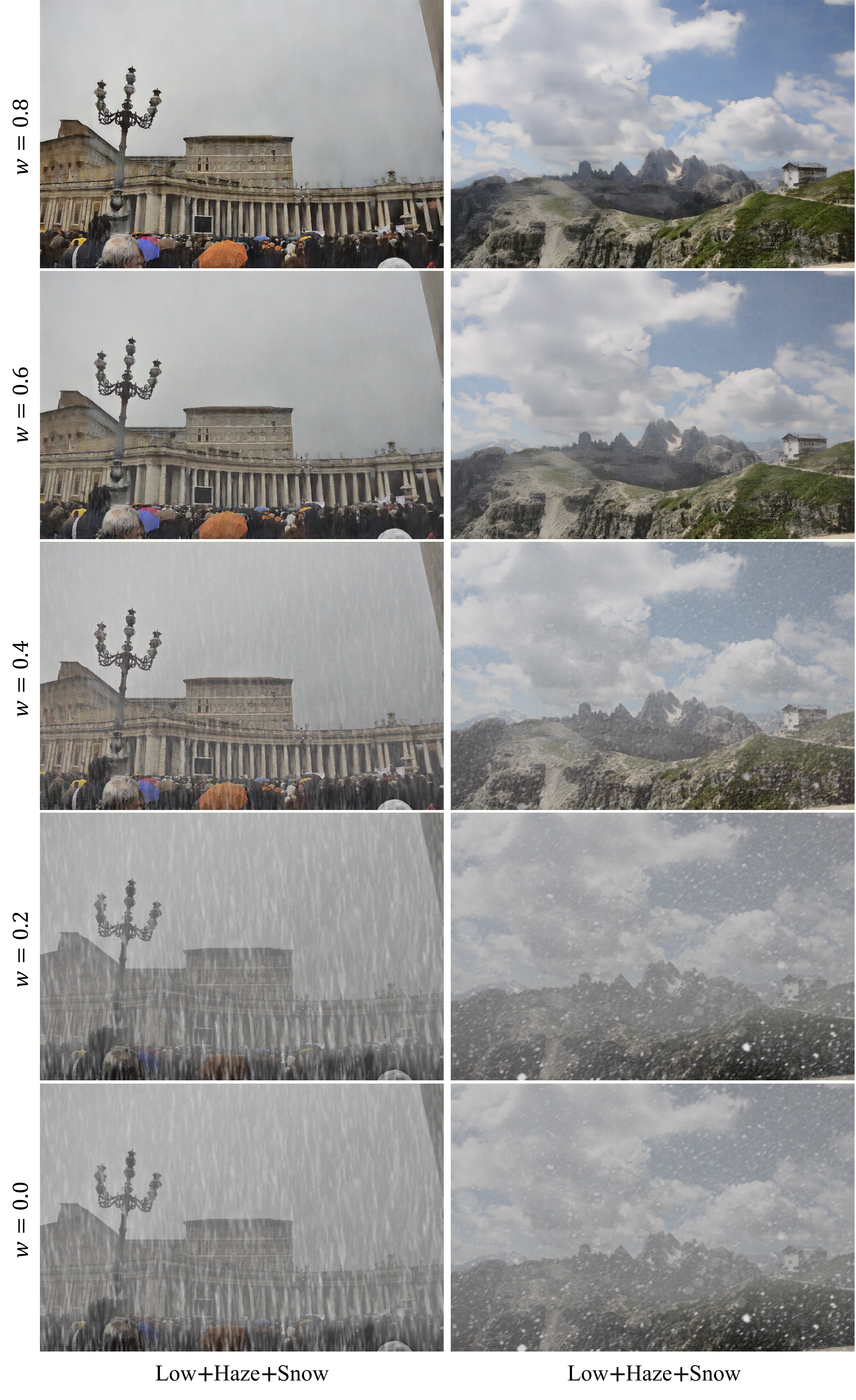}
        \end{tabular}
    \end{center}
    \caption{Qualitative results of Ratio Control Embedding with varying $w$ values for Low$+$Haze$+$Snow and Low$+$Haze$+$Rain.}
    \label{fig:ratio-2}
\end{figure}

\begin{figure}[t]
    \begin{center}
        \def\arraystretch{0.4}
        \begin{tabular}{@{}c}    
            \includegraphics[width=0.9\linewidth]{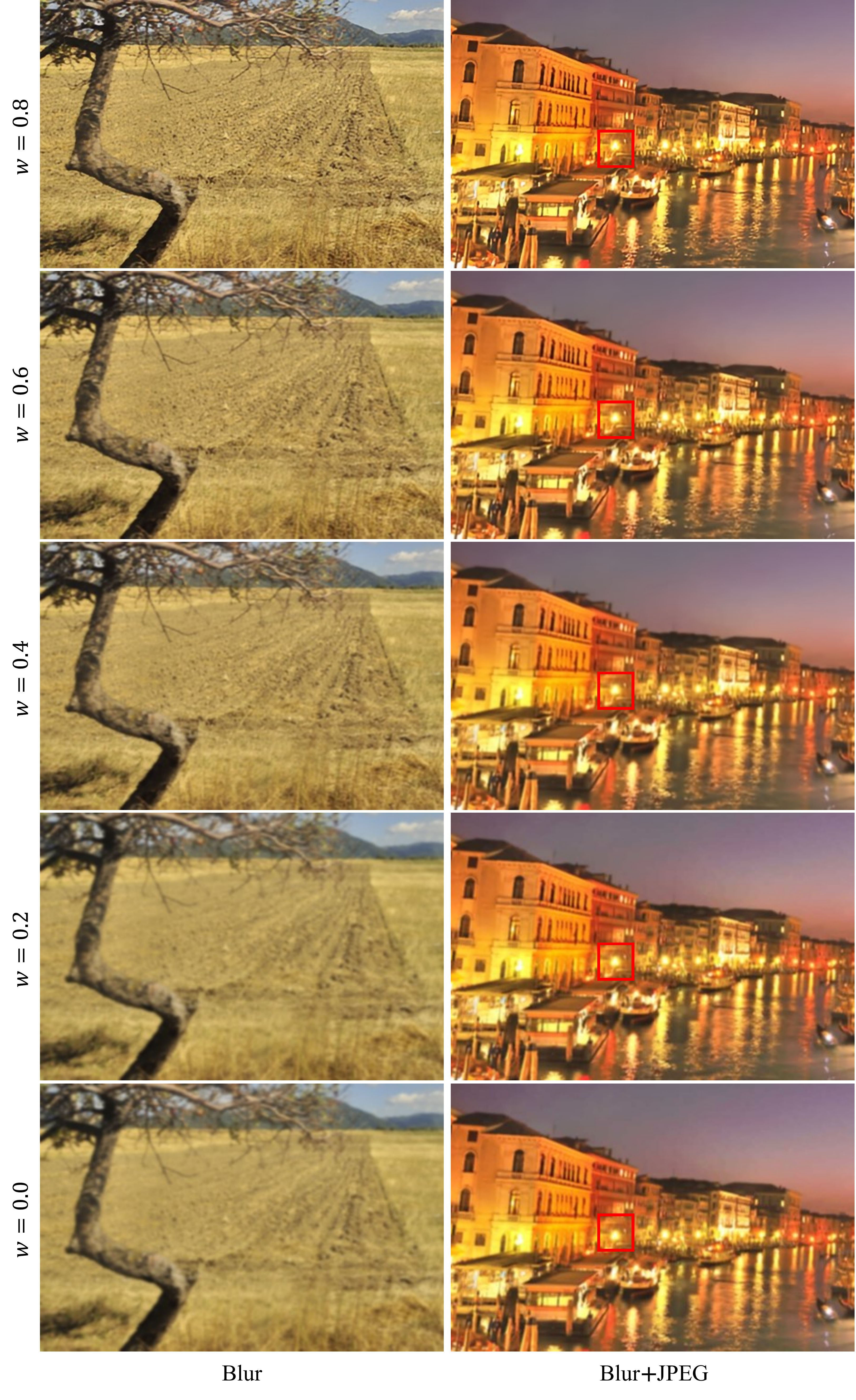}
        \end{tabular}
    \end{center}
    \caption{Qualitative results of Ratio Control Embedding with varying $w$ values for Blur and Blur$+$JPEG.}
    \label{fig:ratio-3}
\end{figure}

With the proposed ratio control embedding and ratio control loss, our model is able to control the intensity of restoration. 
We present qualitative results that demonstrate ratio control across various types of degradation.
By definition, $w=0.0$ corresponds to the identity embedding and produces an output similar to the degraded input image, whereas $w=1.0$ yields a fully restored image.
As shown in Fig.~\ref{fig:ratio-1}, all composite degradations are effectively controlled in a linear fashion as $w$ varies.
Fig.~\ref{fig:ratio-2} presents qualitative ratio control restoration results for three composite degradation types that are excluded from ratio control loss training due to feasibility constraints, yet still demonstrate effective performance.
Fig.~\ref{fig:ratio-3} shows qualitative results for the same experiment conducted on the Blur-Noise-JPEG dataset.
In these cases, as $w$ decreases, the outputs gradually approach the identity operation.
Notably, our method demonstrates reliable linear control over the entire range of $w$ from 0 to 1, despite being trained only on half-degradation without explicit supervision for every possible value of $w$.
\subsection{Selective Restoration}
\begin{figure}[t]
    \begin{center}
        \def\arraystretch{0.4}
        \begin{tabular}{@{}c}    
            \includegraphics[width=0.9\linewidth]{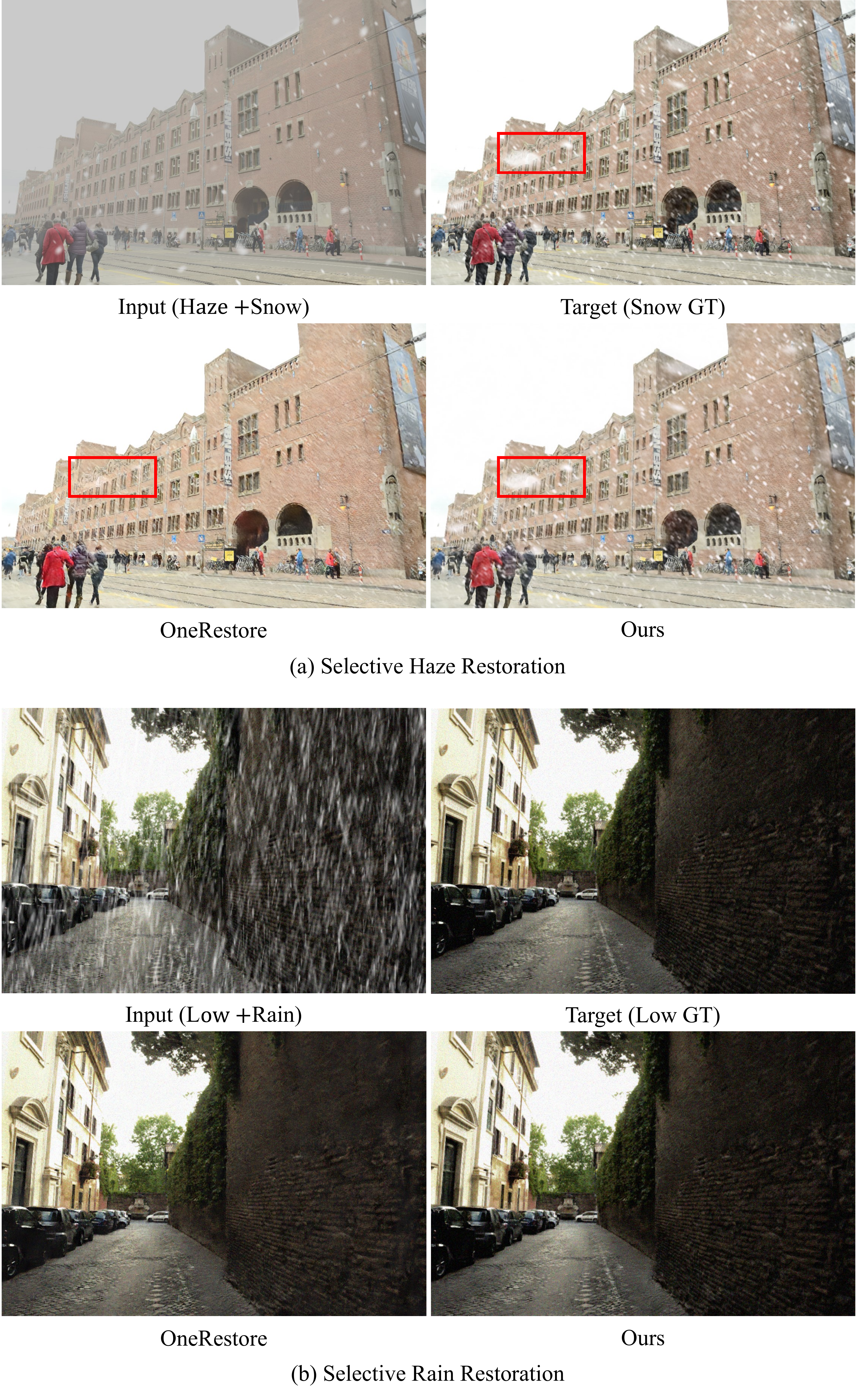}
        \end{tabular}
    \end{center}
    \caption{(a) Qualitative results of selective restoration for the Haze component in Haze$+$Snow images. (b) Qualitative results of selective restoration for the Rain component in Low$+$Rain images.}
    \label{fig:selective-1}
\end{figure}

\begin{figure}[t]
    \begin{center}
        \def\arraystretch{0.4}
        \begin{tabular}{@{}c}    
            \includegraphics[width=0.9\linewidth]{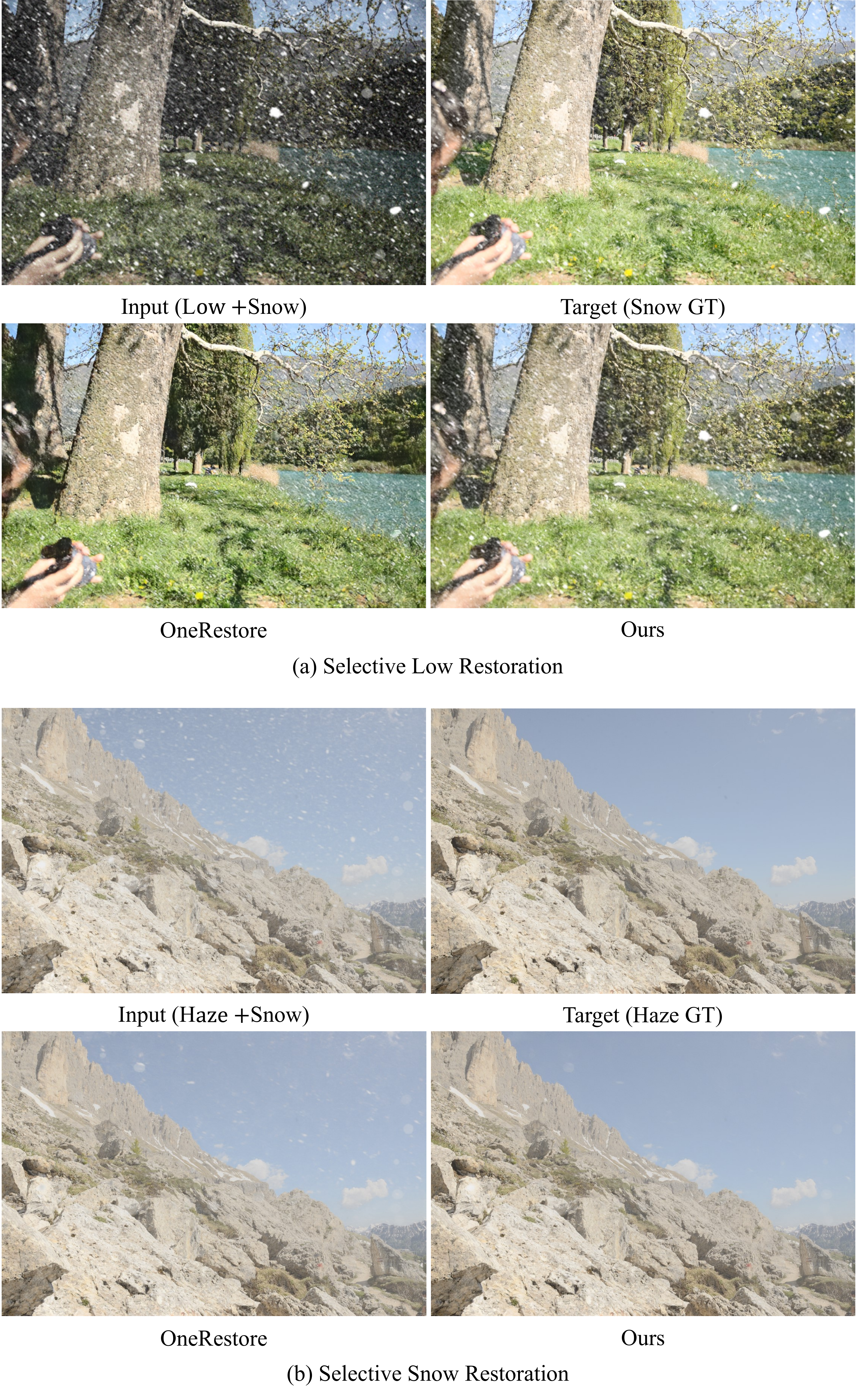}
        \end{tabular}
    \end{center}
    \caption{(a) Qualitative results of selective restoration for the Low component in Low$+$Snow images. (b) Qualitative results of selective restoration for the Snow component in Haze$+$Snow images.}
    \label{fig:selective-2}
\end{figure}

\begin{figure}[t]
    \begin{center}
        \def\arraystretch{0.4}
        \begin{tabular}{@{}c}    
            \includegraphics[width=0.9\linewidth]{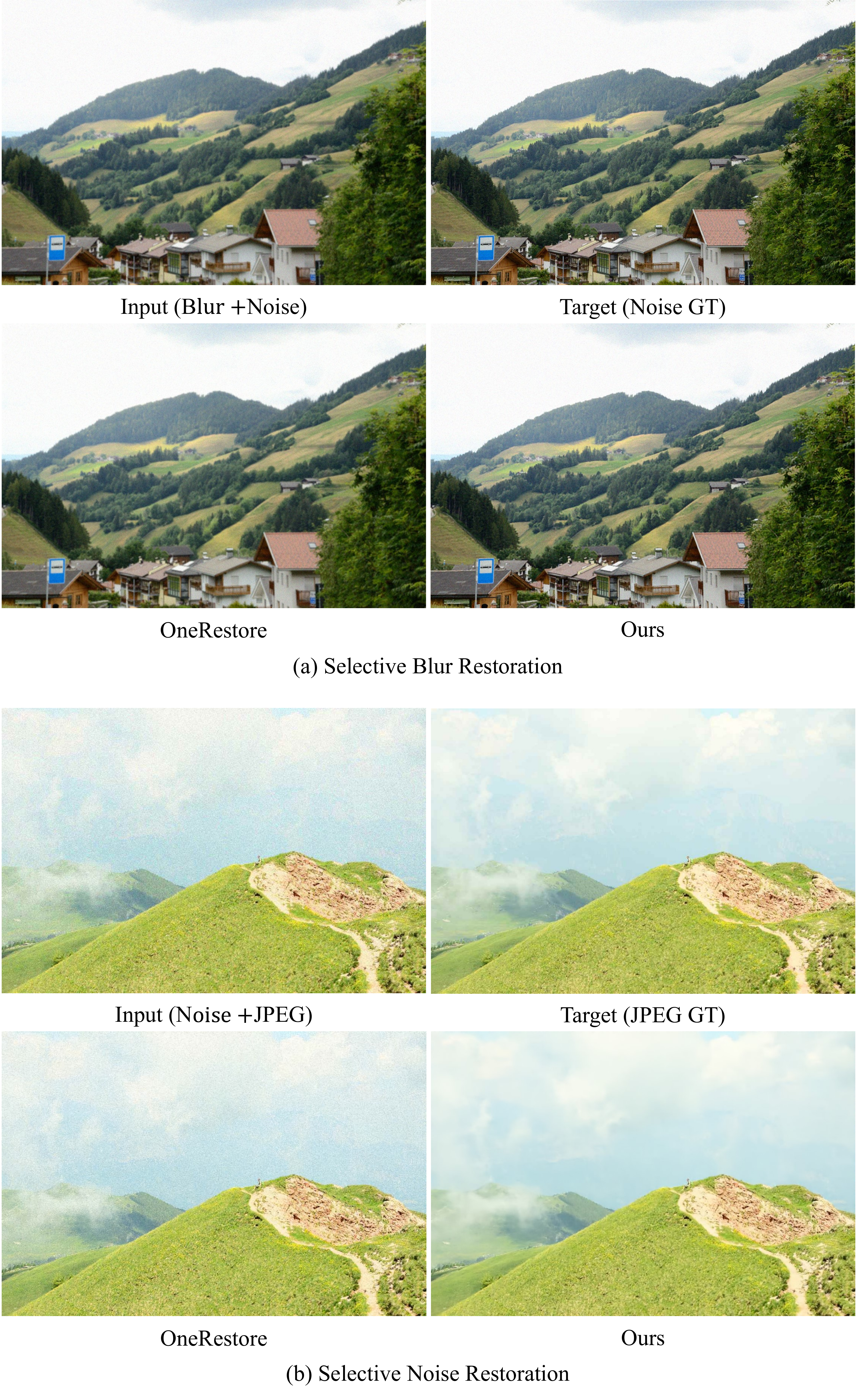}
        \end{tabular}
    \end{center}
    \caption{(a) Qualitative results of selective restoration for the Low component in Blur$+$Noise images. (b) Qualitative results of selective restoration for the Snow component in Noise$+$JPEG images.}
    \label{fig:selective-3}
\end{figure}

\begin{figure}[t]
    \begin{center}
        \def\arraystretch{0.4}
        \begin{tabular}{@{}c}    
            \includegraphics[width=0.9\linewidth]{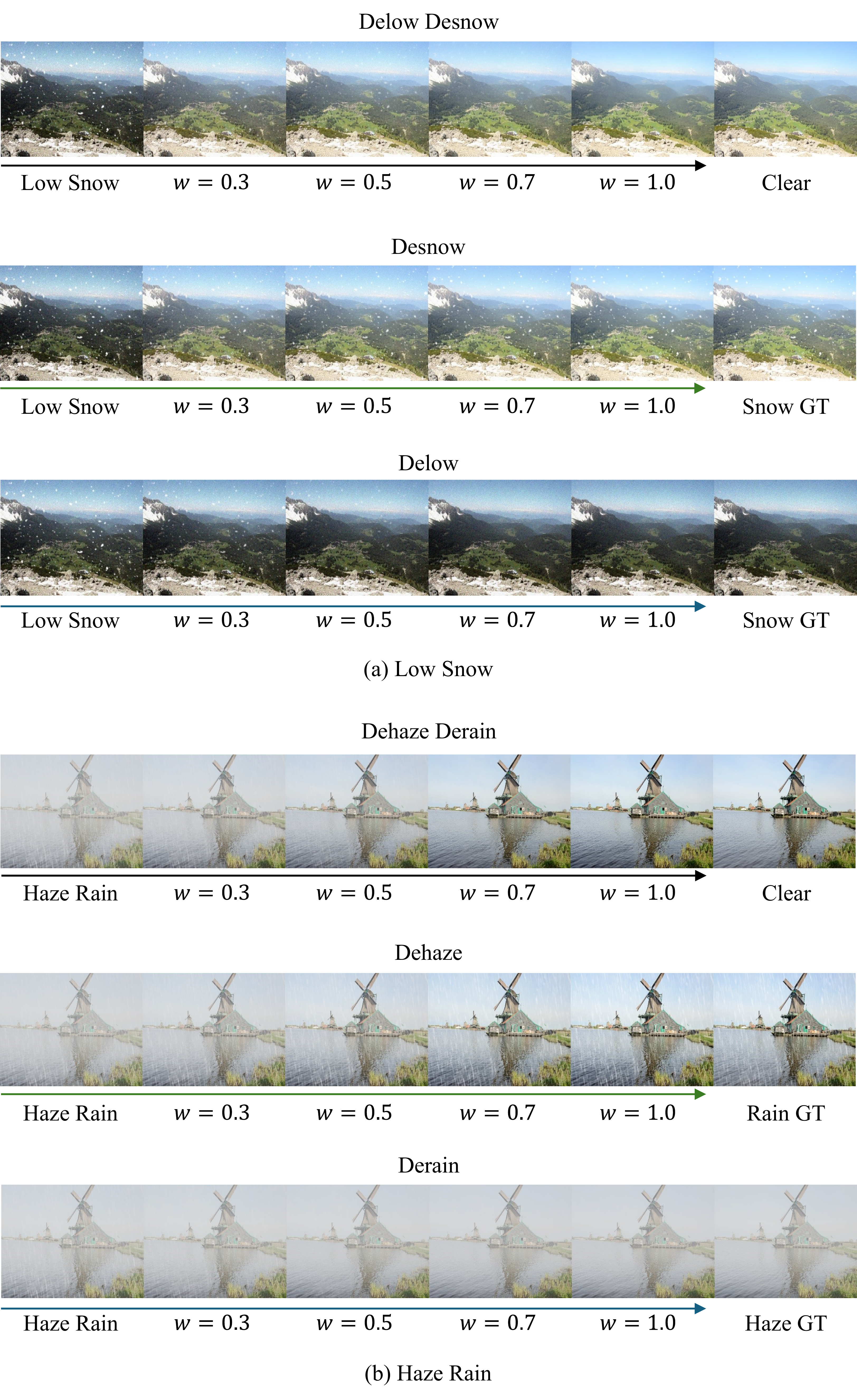}
        \end{tabular}
    \end{center}
    \caption{Example results of simultaneous Ratio Control Restoration and Selective Restoration. For each image, different text embeddings and ratio control values are applied, as indicated by the color of each arrow.}
    \label{fig:ratio-selective}
\end{figure}

Fig.~\ref{fig:selective-1} and Fig.~\ref{fig:selective-2} present qualitative results for selective restoration, where our model removes only the target degradation from composite images while preserving the others.
Fig.~\ref{fig:selective-3} shows qualitative results for the same experiment conducted on the Blur-Noise-JPEG dataset.
These results highlight the ability of our CURE to selectively restore specific degradations while leaving the remaining degradations intact, in contrast to OneRestore, which struggles to provide such selective control.
In Fig.~\ref{fig:ratio-selective}, we further demonstrate that our model can simultaneously perform ratio control restoration and selective restoration on the same image, as indicated by the color of each arrow. 
Remarkably, this capability emerges even though the model was not explicitly supervised for such combined operations during training.
These results underscore the fine-grained controllability of our approach in a variety of composite degradation scenarios.
\subsection{Restoration Order dependency}
\begin{figure}[t]
    \begin{center}
        \def\arraystretch{0.4}
        \begin{tabular}{@{}c}    
            \includegraphics[width=0.9\linewidth]{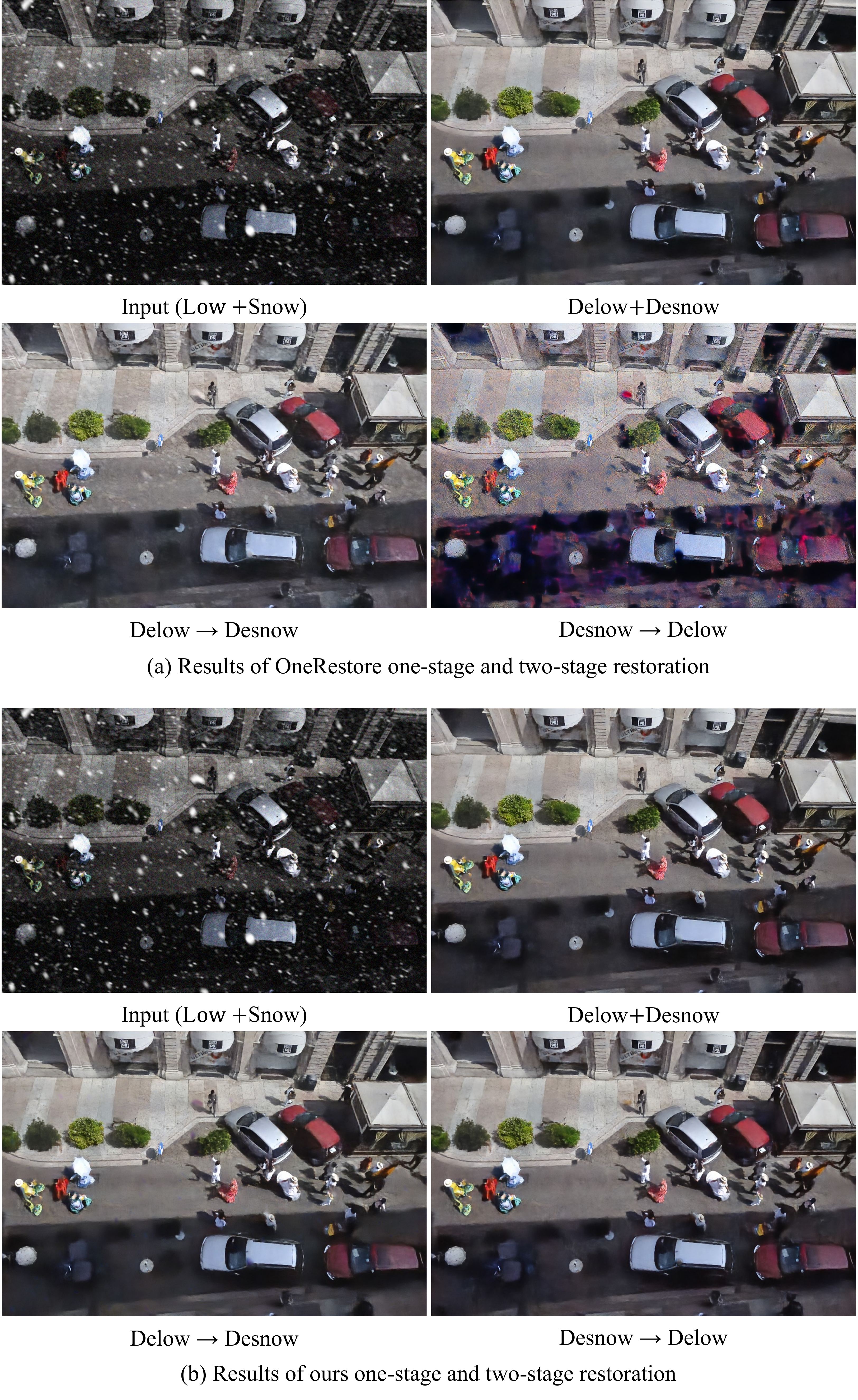}
        \end{tabular}
    \end{center}
    \caption{Comparison of one-stage and two-stage restoration results for (a) OneRestore and (b) Ours on Low$+$Snow images, illustrating the differences between single-stage and two-stage restoration approaches.}
    \label{fig:order-1}
\end{figure}

\begin{figure}[t]
    \begin{center}
        \def\arraystretch{0.4}
        \begin{tabular}{@{}c}    
            \includegraphics[width=0.9\linewidth]{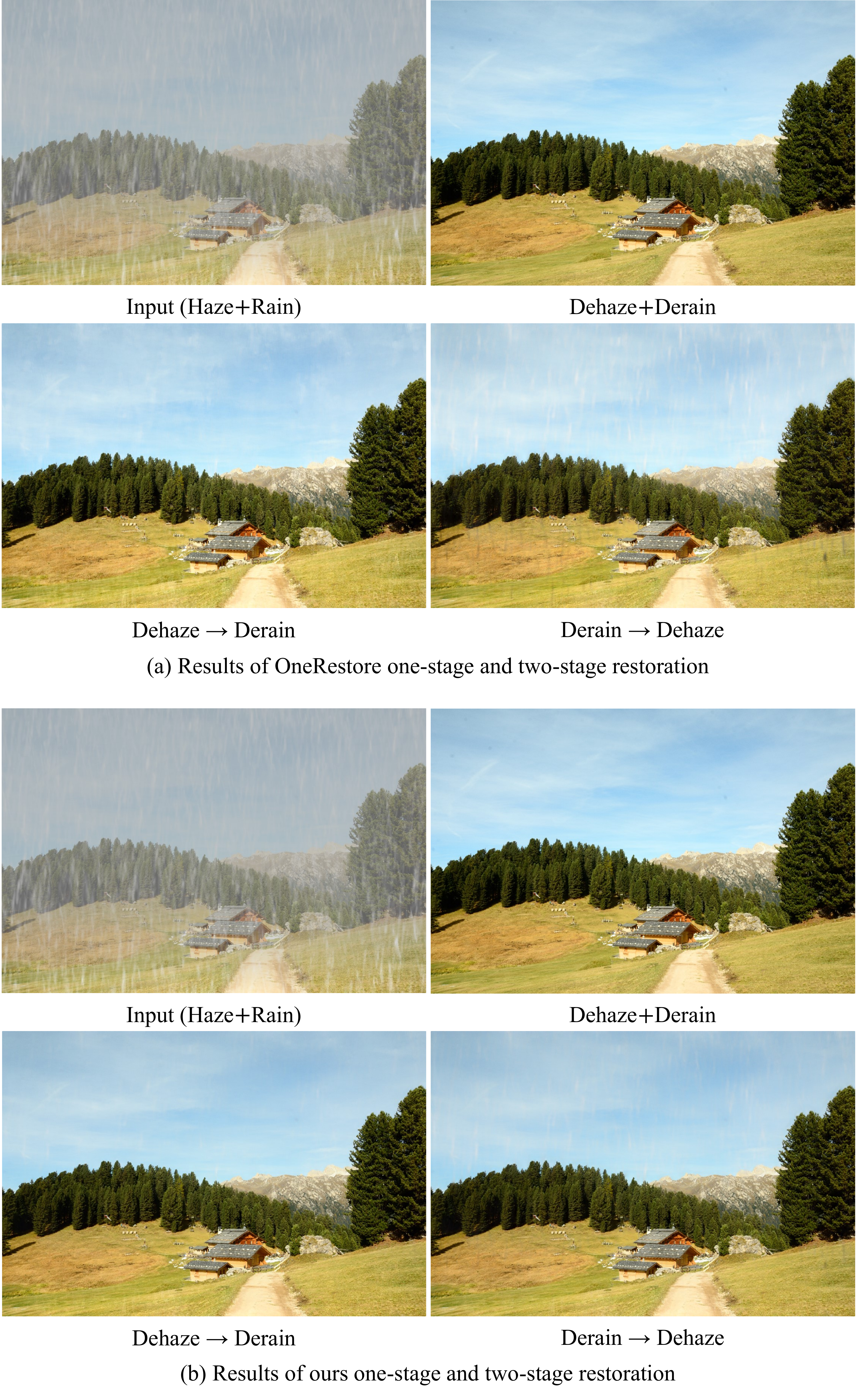}
        \end{tabular}
    \end{center}
    \caption{Comparison of one-stage and two-stage restoration results for (a) OneRestore and (b) Ours on Haze$+$Rain images, illustrating the differences between single-stage and two-stage restoration approaches.}
    \label{fig:order-2}
\end{figure}

Fig.~\ref{fig:order-1} and Fig.~\ref{fig:order-2} highlight the restoration order dependency observed in previous methods and demonstrate that our approach effectively mitigates this issue in composite degradation scenarios.
As shown in these examples, our CURE produces more consistent outputs regardless of the order in which degradations are removed, demonstrating strong order-invariant behavior.
In contrast, OneRestore exhibits noticeable variations depending on the restoration sequence. 
These results visually confirm the effectiveness of our approach in disentangling and restoring individual degradations, regardless of the order in which they are removed.
Furthermore, Fig.~\ref{fig:order-2} presents qualitative results for the same experiment conducted on the Blur-Noise-JPEG dataset.
\subsection{Image Restoration}
\begin{figure}[t]
    \begin{center}
        \def\arraystretch{0.4}
        \begin{tabular}{@{}c}    
            \includegraphics[width=0.9\linewidth]{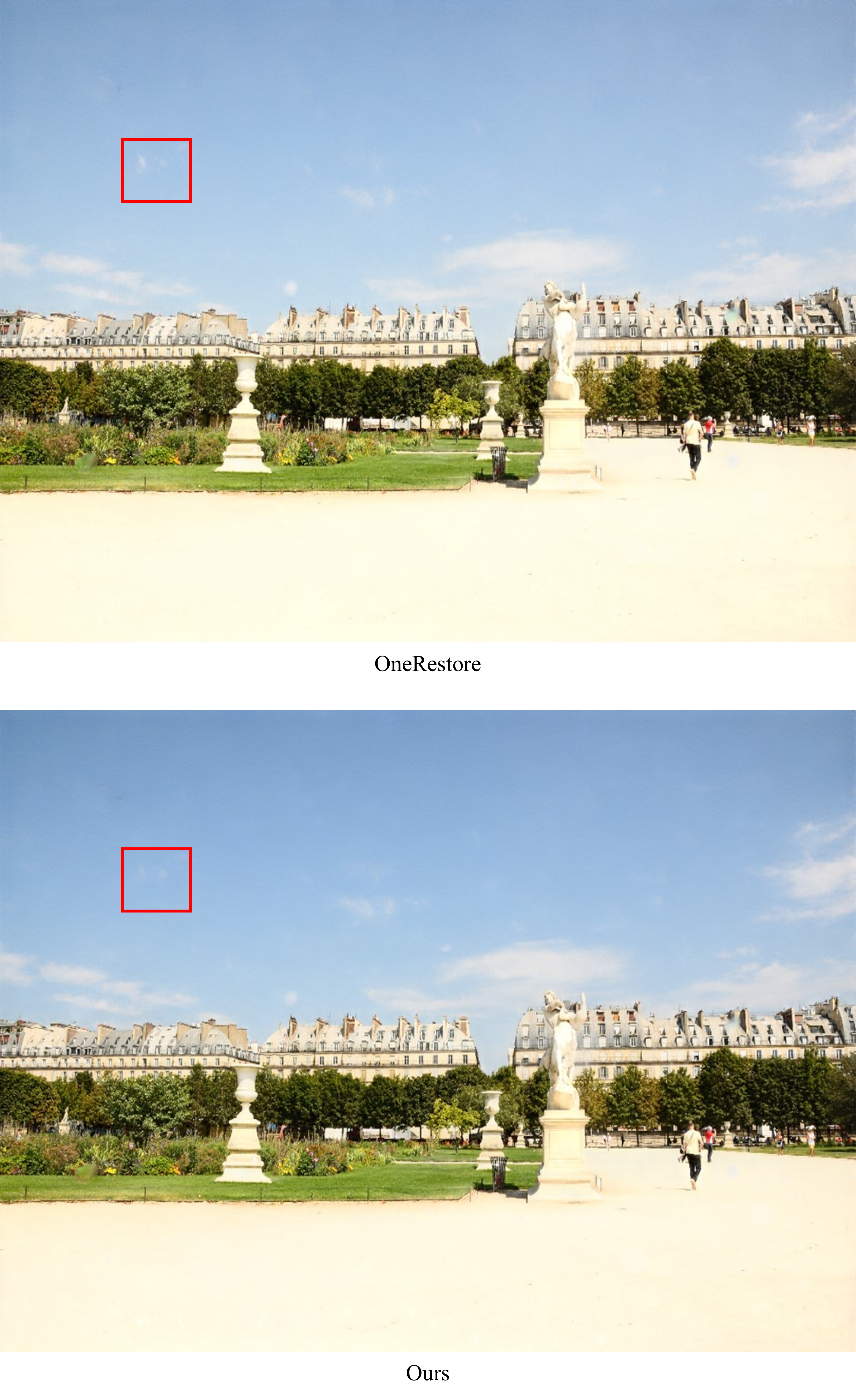}
        \end{tabular}
    \end{center}
    \caption{Qualitative comparison of Haze$+$Snow image restoration results between OneRestore and Ours.}
    \label{fig:restoration-1}
\end{figure}

\begin{figure}[t]
    \begin{center}
        \def\arraystretch{0.4}
        \begin{tabular}{@{}c}    
            \includegraphics[width=0.9\linewidth]{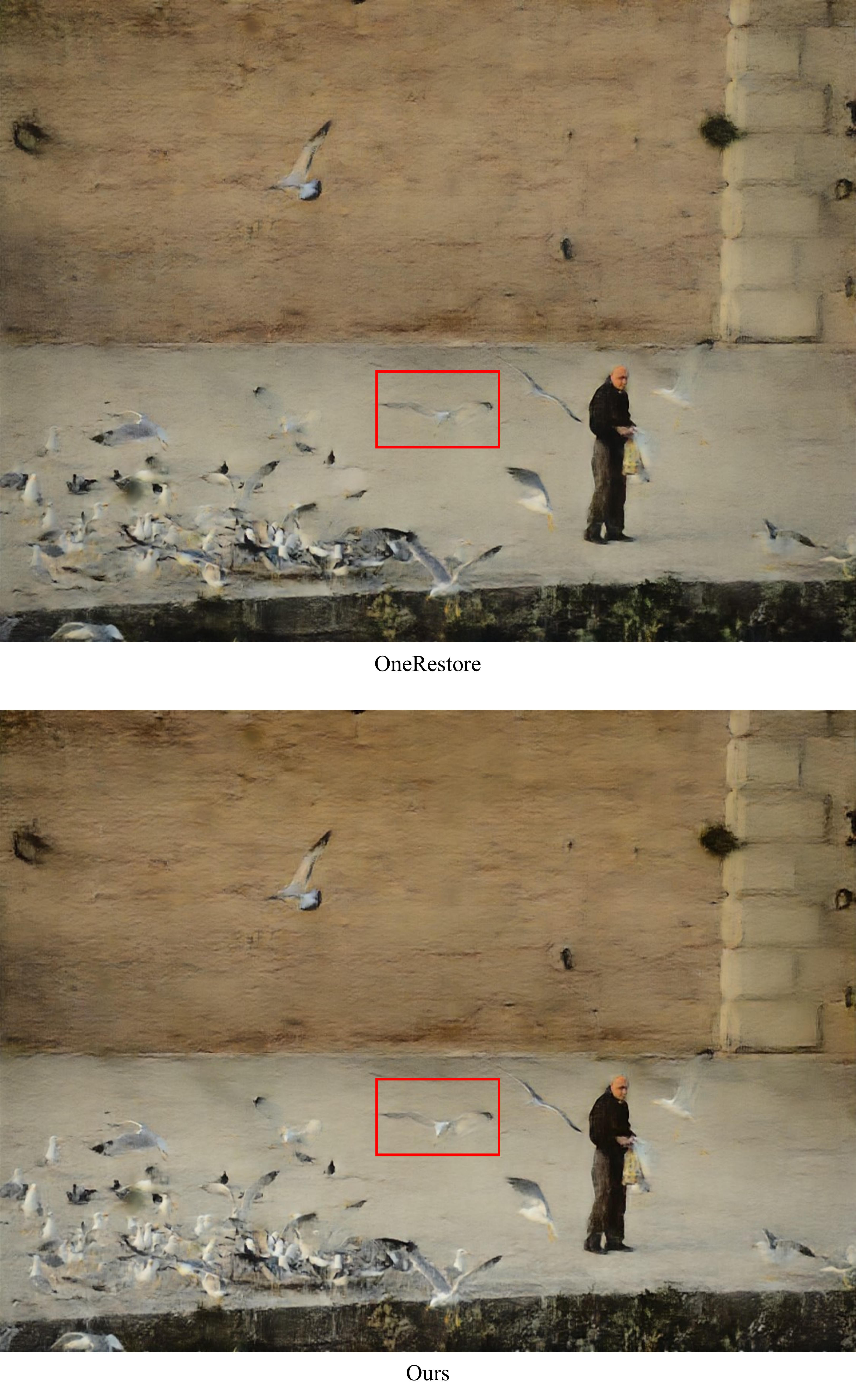}
        \end{tabular}
    \end{center}
    \caption{Qualitative comparison of Low$+$Haze$+$Snow image restoration results between OneRestore and Ours.}
    \label{fig:restoration-2}
\end{figure}

\begin{figure}[t]
    \begin{center}
        \def\arraystretch{0.4}
        \begin{tabular}{@{}c}    
            \includegraphics[width=0.9\linewidth]{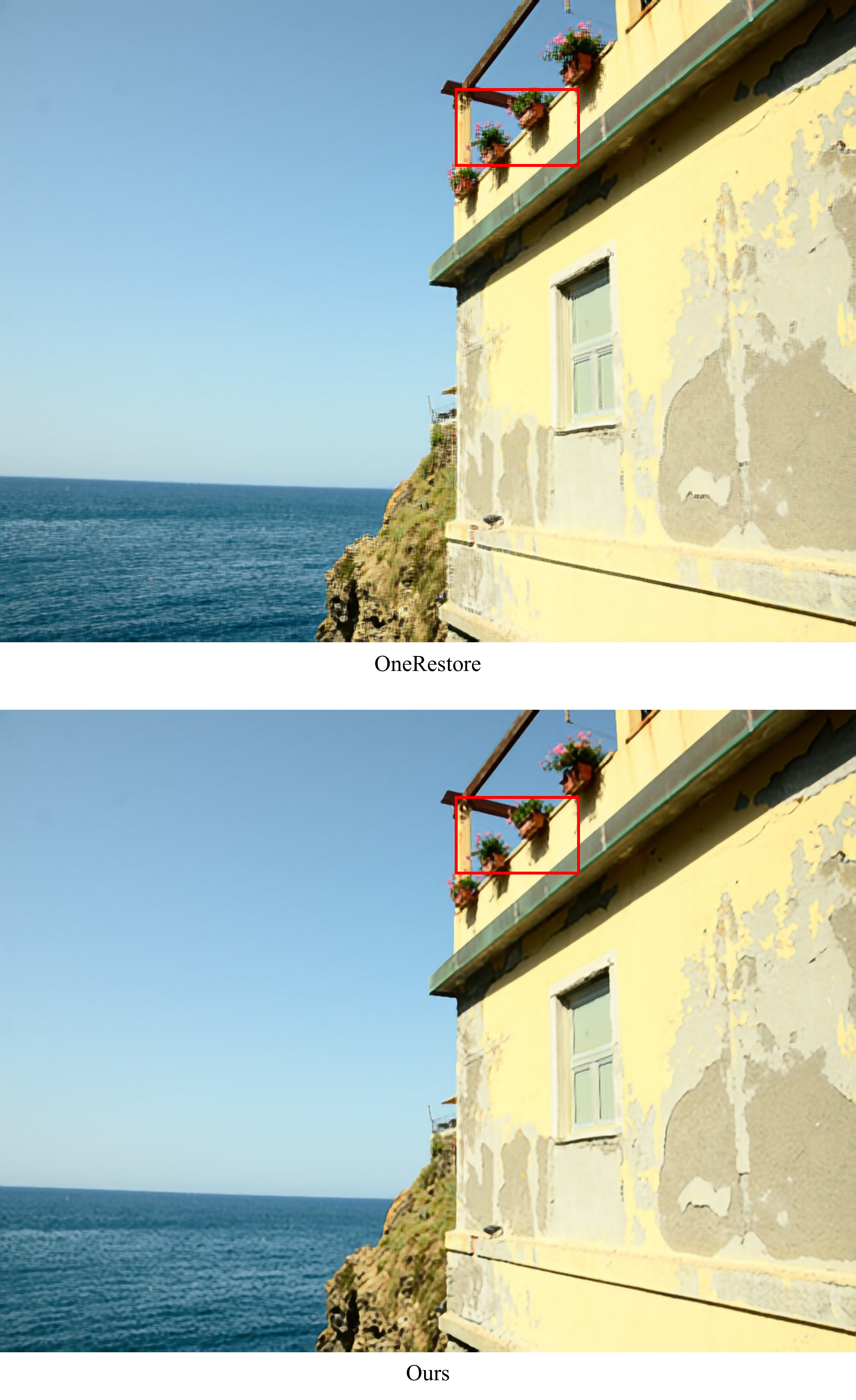}
        \end{tabular}
    \end{center}
    \caption{Qualitative comparison of Blur image restoration results between OneRestore and Ours.}
    \label{fig:restoration-3}
\end{figure}

We present qualitative results for general image restoration using our proposed CCDD-11 dataset.
As discussed in the main paper, our CURE achieves higher quantitative performance compared to previous approaches. 
In addition, qualitative results for $haze+snow$ restoration show that our CURE removes the Snow component more effectively, as highlighted by the red box in Fig.~\ref{fig:restoration-1}.
Fig.~\ref{fig:restoration-2} presents the restoration of $low+haze+snow$ images. 
The red box highlights a challenging region where OneRestore struggles to produce satisfactory results, whereas our CURE achieves noticeably better restoration performance.
In Fig.~\ref{fig:restoration-3}, we show qualitative results for the restoration of Blur images.
In the red box region, it can be seen that our CURE produces better restoration quality compared to OneRestore.
Our proposed method achieves performance improvements even without additional training on three composite images.
We believe this indirectly suggests that our disentangle learning approach helps the model develop a deeper understanding of the distinct properties of each degradation, as well as composite degradations.
\begin{figure}[t]
    \begin{center}
        \def\arraystretch{0.4}
        \begin{tabular}{@{}c}    
            \includegraphics[width=0.9\linewidth]{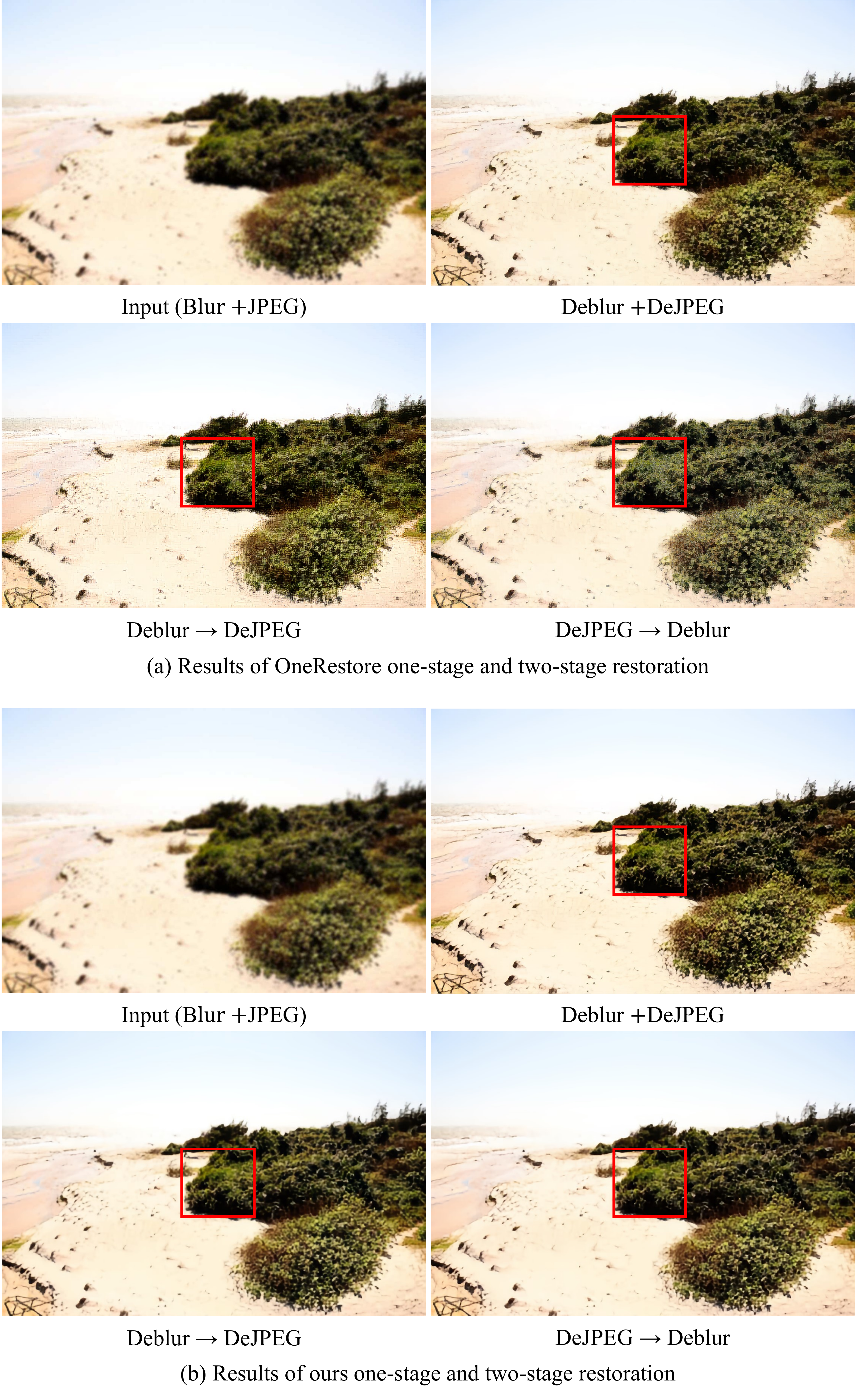}
        \end{tabular}
    \end{center}
    \caption{Comparison of one-stage and two-stage restoration results for (a) OneRestore and (b) Ours on Blur$+$JPEG images, illustrating the differences between single-stage and two-stage restoration approaches.}
    \label{fig:order-3}
\end{figure}

\section{Large Language Model Usage}
\label{sec:llm_usage}
Following the conference rules about using Large Language Models (LLMs), we report how we used LLMs while writing this paper. 
LLMs were used only as basic writing help tools and did not help with research ideas, method creation, experiment planning, or data analysis.
Specifically, LLMs were used for the following purposes:
\begin{itemize}
    \item \textbf{Grammar and Style Refinement}: Improving grammar, sentence structure, and ensuring consistency in academic writing style throughout the manuscript.
    \item \textbf{Logical Structure Enhancement}: Reorganizing sentence flow and improving the logical coherence of paragraphs to enhance readability.
\end{itemize}
All research concepts, methodological innovations, experimental designs, data collection, analysis, and scientific conclusions presented in this work are the result of the original research efforts of the authors. 
The core contributions, including the disentangled prompt learning framework, the four novel loss functions, and the experimental validation, were conceived and developed without the assistance of LLMs.
The authors retain full responsibility for all technical content, claims, and conclusions presented in this paper.
%


\clearpage
{
    \small
    \bibliographystyle{splncs04}
    \bibliography{main}
}

\end{document}